\documentclass[a4,fleqn]{article} 
\usepackage{times}  
\usepackage{helvet}  
\usepackage{courier}  
\usepackage{url}  
\usepackage{graphicx}  
\usepackage{latexsym} 
\usepackage{amsmath} 
\usepackage{amssymb} 
\usepackage{comment} 
\usepackage{multirow} 
\usepackage{amsthm}
\usepackage{bm}

{\hfill \rule[0.3ex]{1ex}{1ex} \par \addvspace{\bigskipamount}}

\newenvironment{proofsk}{\noindent {\bf Proof sketch:}}%
{\hfill \rule[0.3ex]{1ex}{1ex} \par \addvspace{\medskipamount}}

\newcommand{\ie}{i.e.,~}
\newcommand{\eg}{e.g.,~}
\newcommand{\Per}{{ \cal P}}

\newtheorem{definition}{Definition}
\newtheorem{theorem}{Theorem}
\newtheorem{lemma}[theorem]{Lemma}

\DeclareMathAlphabet{\mathpzc}{OT1}{pzc}{m}{it}

\title{Cable Tree Wiring - Benchmarking Solvers on a Real-World Scheduling Problem
with a Variety of Precedence Constraints}
	
\author{Jana Koehler, Joseph B\"urgler, Urs Fontana, Etienne Fux, Florian Herzog,\\  
Marc Pouly, Sophia Saller, Anastasia Salyaeva, Peter Scheiblechner, Kai Waelti\\
contact email: jana.koehler@dfki.de}

\date{Preprint Version July 1, 2020}

\begin{document}

\maketitle

\begin{abstract}
Cable trees are used in industrial products to transmit energy and information between different product parts. To this date, they are mostly assembled by humans and only few automated manufacturing solutions exist using complex robotic machines. For these machines, the wiring plan has to be translated into a wiring sequence of cable plugging operations to be followed by the machine.
 
In this paper, we study and formalize the problem of deriving the optimal wiring sequence for a given layout of a cable tree. We summarize our investigations to model this cable tree wiring problem (CTW) as a traveling salesman problem with atomic, soft atomic, and disjunctive precedence constraints as well as tour-dependent edge costs such that it can be solved by state-of-the-art constraint programming (CP), Optimization Modulo Theories (OMT), and mixed-integer programming (MIP) solvers. It is further shown, how the CTW problem can be viewed as a soft version of the coupled tasks scheduling problem. We discuss various modeling variants for the problem, prove its NP-hardness, and empirically compare CP, OMT, and MIP solvers on a benchmark set of 278 instances. The complete benchmark set with all models and instance data is available on github and is accepted for inclusion in the MiniZinc challenge 2020. 
\end{abstract}

\section{Introduction}

Cable trees are widely used in industrial products to transmit energy and information between different product parts. For example, cable trees are needed  in cars to automate many previously mechanical functions such as moving seats or opening windows and to add new functions such as a voice-controlled navigation or an onboard entertainment system. It is thus not surprising that for example a car like the VW Golf~7 contains 14 cable trees with a total of 1633 cables.  
 
The manufacturing of cable trees usually relies on cheap manual labour performed in low-cost countries where humans plug cables into harnesses following a wiring plan. Only few automated manufacturing solutions exist, which rely on complex robotic machines. These machines execute a sequence of wiring operations that highly qualified technicians develop by analyzing the wiring plan. With the continuing tendency towards customer-specific and resource-efficient just-in-time manufacturing, smaller batch sizes of cable trees need to be manufactured requiring a frequent change of wiring plans, for which wiring sequences should be derived instantly. Scaling up human expertise to such frequent changes is simply impossible, which explains a growing interest in the intelligent automated manufacturing of cable trees. This interest is also nourished by a further miniaturization of cable harnesses, which will make their manual manufacturing impossible.

In this paper, we study the problem of automatically computing the wiring sequence to manufacture a cable tree and formalize it as the problem of cable tree wiring CTW.\footnote{Our problem is motivated by the automatic manufacturing of cable trees on a Zeta machine by Komax AG, Switzerland. This machine is a highly sophisticated robot that prepares and cuts cables, adds contacts, and then inserts the cables into cable harnesses mounted on a palette. A video of such a machine wiring a cable tree can be watched at https://www.youtube.com/watch?v=cvfnb0thjXA.}  To wire a cable tree on this machine, two problems have to be solved by experienced human technicians today: 
  
\begin{enumerate}
\item Layout: At which positions do cable harnesses have to be mounted on the palette such that the robot arm can insert cables into the designated harness cavities?
  	
\item Insertion order: Given a layout of harnesses on the palette, in which order are the cable ends to be inserted into harness cavities such that a robot arm of the machine can fast and safely produce the desired wiring?
\end{enumerate}

\begin{figure}[htb]
	\begin{center} 
		\includegraphics[width=.8\textwidth]{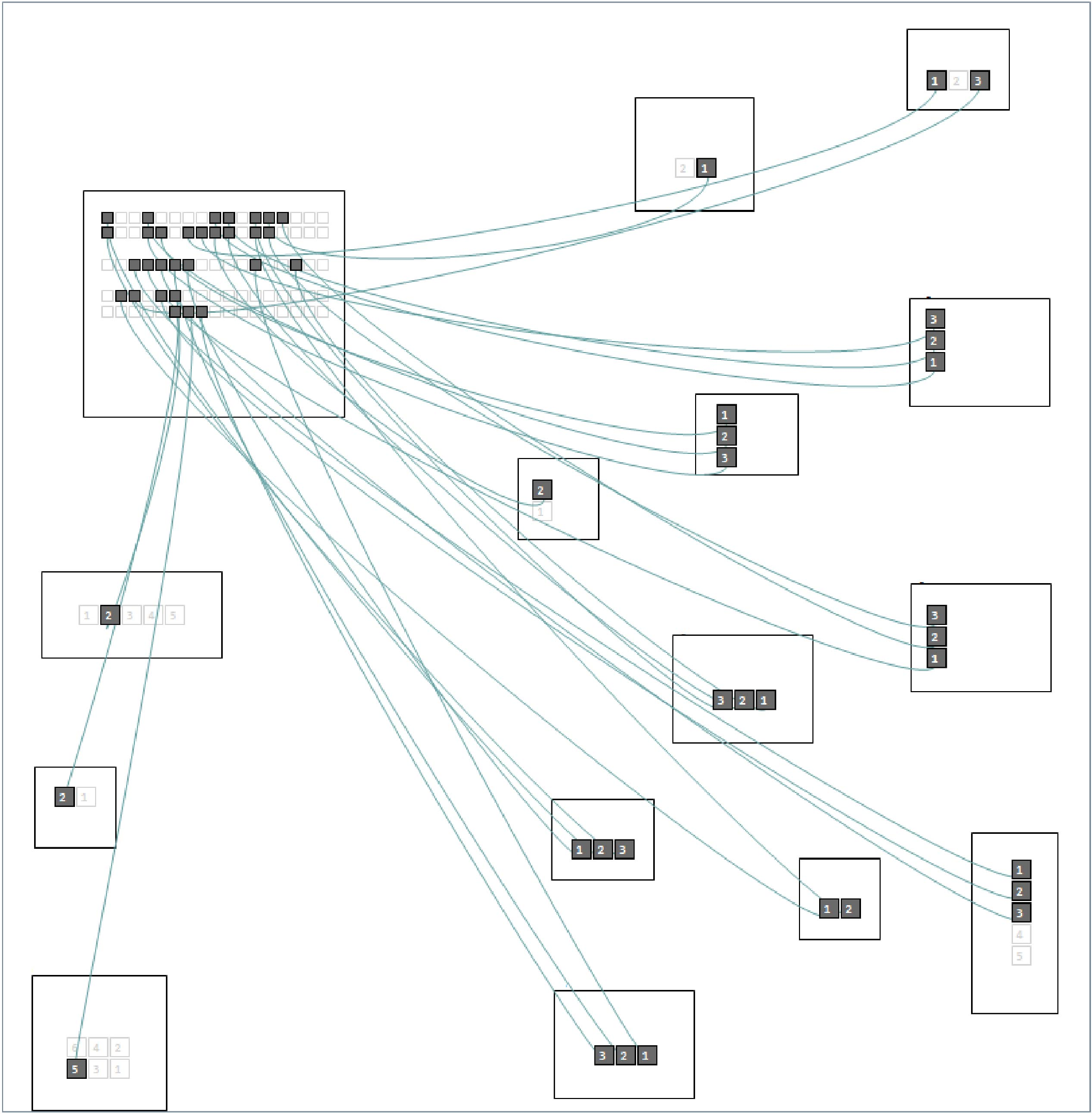}
		\caption{\label{fig:cabletree} Layout of a cable tree on a palette and desired wiring.} 
	\end{center}  
\end{figure}

Figure~\ref{fig:cabletree} illustrates the two problems. The large rectangle represents the palette surface. On the palette, we see a number of small rectangles that represent the positions of the cable harnesses (the layout) on the mounting palette. Each rectangle contains several light and dark grey numbered rectangular fields. These are the harness  cavities into which the cables need to be inserted. Dark grey cavities are filled with cables, whereas light grey cavities are left unused. These unused cavities might be required for other variants of a cable tree. 

Note that the figure shows a 2D-model of the palette, harnesses, and the desired wiring of cavities abstracting from 3D-geometric information. In reality, cable harnesses and cavities can occur in many different shapes (rectangular, oval, or round). The wiring of the cable tree is illustrated by Bezier curves showing the connections between cavities. Note that these curves do not represent the real geometric dimension nor physical behavior of the cables---cables can be up to several meters long and can have very different diameters and physical properties. 

The work in this paper focuses on solving the insertion order problem and assumes that a fixed layout of harnesses on the palette is given. Based on such a layout, we are looking for an enumeration of the dark grey cavities---a robust and fast sequential order (a permutation) of insertion operations that plug all cables into their designated cavities. Constraints restrict the positions that cavities can take in the permutation and are derived from an analysis of the cable behavior and properties of the machine. For example, a constraint can express that a cable end needs to be inserted into a cavity $A$ before another cable end is inserted into a cavity $B$ as otherwise the robot arm of the machine might not be able to approach cavity $B$, which can be occluded by the cable plugged into $A$. The computation of these constraints is beyond the scope of this paper.

The paper is organized as follows: Section~\ref{related} reviews related work to position the problem. Section~\ref{permutation} formalizes the CTW problem, proves its NP-hardness and identifies subclasses of the problem that can be solved in polynomial time.  Section~\ref{benchmark} introduces the CTW benchmark set of 278 real-world and artificial instances. Section~\ref{models} summarizes modeling variants for the CTW problem and presents a comprehensive tool chain supporting a benchmarking of constraint-, mixed-integer, and optimization modulo theories solvers. In Section~\ref{solvers}, we discuss our empirical findings when testing the following solvers:

\begin{itemize}
	\item CP-SAT solvers
	\begin{itemize}
		\item IBM Cplex CP Optimizer 12.10 (C$\sharp$ API)~\cite{cplex}
		\item Google OR-Tools CP-SAT Solver 7.5.7466 (Windows Executable)~\cite{ortools}
		\item Chuffed 0.10.3 (Windows Batch File/Executable)~\cite{chuffed}
	\end{itemize} 
	\item MIP solvers
	\begin{itemize}
		\item IBM Cplex MIP solver 12.10~(C$\sharp$ API)\cite{cplex} 
		\item Gurobi 9.0.1 (C$\sharp$ API)~\cite{gurobi}
	\end{itemize} 
	\item OMT solvers
	\begin{itemize}
		\item Microsoft Z3 4.8.7 (Windows Executable)~\cite{z3} 
		\item OptiMathSAT 1.5.1 (C API)~\cite{optimathsat}
	\end{itemize} 
\end{itemize}

Section~\ref{conc} concludes the paper with a discussion of interesting open research problems.

\section{Related Work}
\label{related}

Permutations represent an abstract characterization of many manufacturing problems where an optimal sequential ordering of a given set of manufacturing steps has to be computed. They have for example been subject to intense research in the context of routing, scheduling, or assignment problems~\cite{lageweg,gao,sawada,benoist,wang2013single}. Frequently, boundary conditions lead to various forms of constraints, notably precedence constraints, which can also occur as disjunctions and which add additional complexity to a problem formulation~\cite{baptiste,pferschy,abbou}. To optimally solve such a problem, a constraint satisfaction or mixed-integer programming approach can be followed, but also a number of heuristic approaches have been popular in the literature, \eg greedy or tabu search~\cite{grabowski,ruiz}, ant colony and swarm particle algorithms~\cite{solnon,gottlieb,tasgetiren}, or simulated annealing~\cite{osman}.

In Section~\ref{permutation}, we position our problem as a traveling salesperson problem with (disjunctive) precedence constraints (TSPPC) and time-dependent edge costs (TDTSP)~\cite{vajda,picard, vander}, which depend on the current tour, \ie which other cities (cavities for wiring) have or have not been visited before. This means, CTW belongs to the TDTSPPC class of TSP problems. A related variant of such a problem with time windows is for example studied in~\cite{fagerholt}, variants of vehicle routing problems with time-dependent travel times are studied for example by~\cite{haghani,donati}.  Precedence constraints also occur in many other settings for example in vehicle routing problems, when customer visits have to happen in a specific order or within a time window or when vehicles have to meet. An example for these type of problems is described in~\cite{bredstrom}, however their constraints are quite different from ours. We were not able to find work that exactly addresses the unique combination of precedence constraints and tour-dependent edge costs as it occurs in our problem.

The CTW problem can also be seen as a variant of the coupled task scheduling problem where a set of jobs consisting of two operations with processing times is given, which should be scheduled on a single machine observing a given time-lag~\cite{condotta2012scheduling, orman1997complexity}. In the CTW problem, the coupled tasks represent the two operations that insert the ends of a cable. As we discuss in Section~\ref{permutation}, we wish the two ends of a cable to be plugged right after one another. In contrast to the coupled task scheduling problem, we do not enforce a time-lag smaller than a given constant for any coupled insertion operations, but we incur two different penalties. One counts the number of decoupled insertion operations and the other depends on the amount of lateness by which the two insertion operations are interrupted. This makes our problem a soft version of the coupled task scheduling problem.

Different approaches to the formulation of permutation problems as CSP problems have been systematically studied in~\cite{walsh,smith}, whose results influenced the modeling approach that we present in Section~\ref{models}. The work in~\cite{walsh,smith} also demonstrated that there is no model that is best suited for all problems, which motivated the comparison of different modeling variants that we present in this paper. One of these variants used in the context of MIP solvers is using a so-called big-M reformulation~\cite{camm,ruiz} to effectively rewrite disjunctive constraints. Permutation problems with disjunctive constraints lead to AND/OR constraint graphs for which first analysis techniques have been discussed in~\cite{austrin,mapa}. The impact of very large sets of disjunctive constraints on the difficulty of permutation problems has not been studied very widely, however our experiments show that modern solvers handle problems with disjunctive constraints quite effectively.  An early study of disjunctive precedence constraints is described in~\cite{chen}.

A similar study to ours that empirically compares different MIP solvers on various models for the standard job shop scheduling problem is presented in~\cite{ku}. These models also contain precedence and disjunctive constraints. Whereas the authors use a well-known problem to investigate the progress made by MIP solvers, we are introducing a novel benchmark set that combines two very different and practically relevant TSP variants in an interesting way and which also seems to be the first known representative of the soft coupled task scheduling problem. Furthermore, our empirical analysis spans over three different classes of solvers, for which we developed an elaborate tool chain supporting the conversation of models and data. A paper that compares MIP and CP solvers on the hospitals/residents problem is~\cite{manlove}, however, cross-approach comparisons of solvers seem to be rather infrequent. Our findings in Section~\ref{solvers} demonstrate the recent progress made by CP-SAT solvers and the impact of advances such as clause-direct conflict learning~\cite{zhang} and lazy clause generation~\cite{ohrimenko} leading to an impressive performance of these solvers.

\section{Formalization of the Cable Tree Wiring Problem}
\label{permutation}

The cable tree wiring problem CTW can be considered as a scheduling problem where an optimal schedule for inserting each given cable end into its designated cavity needs to be determined. Note that two different cable ends can never be inserted into the same cavity and one cable end cannot be inserted into two different cavities. As this matching of cable ends to cavities is given initially, the insertion of cable end $i$ in its designated cavity hence uniquely determines a job $c_i$, which we formally define as follows:

\begin{definition}[Job]
A job $c_i$ is defined as the task of inserting a cable end $i$ into its corresponding cavity.
\end{definition}

The CTW problem can be considered as the problem of finding an optimal schedule for executing these jobs while satisfying certain constraints. The schedule is described by the permutation sequence of insertions on a single machine executing the jobs as fast as possible.  Note that the duration time of an insertion operation is not influenced by its position in the permutation, but by the layout, \ie the position of the cavity on the palette, which determines the travel time of the robot arm. For each insertion operation, the robot arm has to pick up the cable from a fixed position in the storage, then travel to the cavity on the palette, plug the cable, and then travel back to the storage to pick up the other cable end or the next cable.\footnote{When computing the layout of the harnesses on the palette, we indeed minimize travel costs for the robot arm using another optimization solution, which is not subject of this paper.} Thus, the travel times are identical for all valid permutation sequences and do not need to be considered in the problem formalization.

We consider two types of cables, which we denote as one-sided and two-sided cables. For two-sided cables, both ends must be inserted into cavities. For one-sided cables only one of their ends needs to be inserted into a cavity. We define the follwing convention for labeling the corresponding jobs:

\begin{definition}[Labeling of jobs, job pair $\langle c_i, c_j\rangle$]
Let $b$ be the number of two-sided cables in a CTW instance. We label the two ends of a two-sided cable  $i$ and $j=b+i$, where $i = 1, \dots, b$. Every two-sided cable hence defines a job pair $\langle c_i, c_j\rangle$ where $i\in \{1, \dots, b\}$ and $j = b + i$. Let $n$ be the number of one-sided cables in a CTW instance labelled with $i = 2b+1, \dots, 2b+n$. The one-sided jobs are hence the jobs $c_i$ where $i = 2b+1, \dots, 2b+n$.
\end{definition}

A solution of a CTW instance with $k=2b+n$ jobs is described by a permutation $\Per$ of length $k$. This permutation sequence is an ordered set in which each job occurs exactly once.

\begin{definition}[Solution $\Per$]
\label{solution}
Let $p(c_i)$ be the position of a job $c_i$ in $\Per$. $\Per$ is a solution of the CTW problem if $p$ is an invertible function from $\{c_i\}_{i\in \{1, \dots, k\}}$ to $\{1, \dots, k\}$ such that
\begin{subequations}
\begin{align}
&\forall x \in \{1, \dots, k\} \,\, \exists\,! \, i \in \{1, \dots, k\}\,\, \text{ s.t. } \,\, p(c_i) = x\\
&\forall i, j \in  \{1, \dots, k\}\,\, \text{with}\,\, i \neq j\, : \,  p(c_i) \neq p(c_j)\text{ ,}
\end{align}
\end{subequations}
where $\exists \, !$ means that there exists a unique one.
\end{definition}

As mentioned in the introduction, the position of a job in the permutation sequence
is restricted by the behavior of cables and the machine. To capture these restrictions, constraints are formulated over the jobs. These constraints occur in three different forms: 

\begin{definition}[Atomic Precedence Constraint, sets ${\cal A}$ and ${\cal A}_s$]
Let $c_i$ and $c_j$ be two jobs in a CTW instance. An atomic precedence constraint
$c_i \triangleleft c_j$ with $i, j  \in \{1, \dots, k\}$ and $i \neq j$ specifies that job  $c_i$ must be executed before job $c_j$. The set of all atomic precedence constraints in a CTW instance is denoted by ${\cal A}$. We further define another set of soft atomic precedence constraints ${\cal A}_s$, which do not necessarily have to be satisfied by a valid solution permutation $\Per$, but will lead to a penalty when violated. The two sets are disjoint, \ie ${\cal A} \cap {\cal A}_s = \emptyset$. A solution $\Per$ satisfies a (soft) atomic precedence constraint $c_i \triangleleft c_j   \in {\cal A} \cup {\cal A}_s$  if and only if $ \Per \vdash p(c_i) < p(c_j)$.
\end{definition}

Given any arbitrary pair of jobs $c_i$ and $c_j$, an atomic precedence constraint $c_i \triangleleft c_j$ or $c_j \triangleleft c_i$ may or may not occur in a problem instance. Sometimes, even both constraints can occur, which renders a problem instance unsatisfiable, because for example the layout of harnesses was chosen in an unfortunate way.

\medskip

Disjunctive precedence constraints $\cal{D}$ combine two atomic precedence constraints in a disjunction and occur in a limited syntactic form in CTW.

\begin{definition}[Disjunctive Precedence Constraint, set $\cal{D}$]
Given a job pair $\langle c_i, c_j\rangle$ of a two-sided cable and a job $c_l$ from another one-sided or two-sided cable, two syntactic forms of disjunctive precedence constraints can occur in a CTW instance: 
 \begin{enumerate}
 	\item ${\cal D}_1: c_l \triangleleft c_i \vee c_l \triangleleft c_j$
 	\item ${\cal D}_2: c_l \triangleleft c_i \vee c_j \triangleleft c_l$ or switching $i$ and $j$:  $c_l \triangleleft c_j \vee c_i \triangleleft c_l$
 \end{enumerate}
For  $l \in \{1, \dots, k\}$ and a job pair  $\langle c_i, c_j\rangle$ with $i \in \{1, \dots, b\}$ and $j = i+b$, a solution $\Per$ satisfies the constraint
\begin{itemize}
\item $(c_l \triangleleft c_i \vee c_l \triangleleft c_j) \in {\cal D}_1\,$ if and only if $\, \Per \vdash p(c_l) < p(c_i)\,$ or $\,\Per \vdash p(c_l) < p(c_j)$
\item $(c_l \triangleleft c_i \vee c_j \triangleleft c_l) \in {\cal D}_2\,$ if and only if $\,\Per \vdash p(c_l) < p(c_i)\,$ or $\,\Per \vdash p(c_j) < p(c_l)$.
\end{itemize}
\end{definition}

Note that  two constraints of the form $c_l \triangleleft c_i \vee c_l \triangleleft c_j$  and $c_i \triangleleft c_l \vee c_l \triangleleft c_j$ ($c_l$ and $c_i$ flipped) can never occur in the set ${\cal D}_1$ for the same problem instance, because a cable has at most two ends, \ie two job pairs $\langle c_i, c_j\rangle$ and $\langle c_l, c_j\rangle$ cannot exist together in the same instance. Furthermore, note that whenever a disjunctive constraint for three jobs $c_i,c_j,c_l$ occurs in  ${\cal D}_1$ with $\langle c_i,c_j\rangle$ being a job pair, a constraint in ${\cal D}_2$ can never be present for the same three jobs and vice versa. For the set ${\cal D}_2$, both variants of the constraint with $c_i$ and $c_j$ being flipped can occur within the same instance.

\medskip
Direct successor constraint formalize coupled tasks in a CTW instance. A Zeta machine can insert the end $i$ of one cable into a cavity, put the other end of the same cable $j$ into storage for later insertion, and continue to work on a new cable. However, some cables are too short for storing and thus require that plugging of the cable must proceed without interruption. Note that direct successor constraints are specific to cavities, not cables. Depending on the position of the cavity on the palette, storage of one end might be possible, but storage of the other end might be not.
 
\begin{definition}[Direct Successor Constraint, set  $\cal{S}$]
Let $\langle c_i,c_j\rangle$ be a job pair. A direct successor constraint  $c_i \blacktriangleleft c_j$ (or $c_j \blacktriangleleft c_i$) formulates a strong atomic precedence constraint, which requires that the cable end $j$ of a two-sided cable is immediately inserted after $i$ is inserted or anytime before $i$ (or $i$ to follow immediately after or anytime before $j$ respectively). A solution $\Per$ satisfies a direct successor constraint $c_i \blacktriangleleft c_j$ for a job pair $\langle c_i,c_j\rangle$ if and only if $\Per \vdash p(c_j) = p(c_i) + 1$ or $\Per \vdash p(c_j) < p(c_i)$.
\end{definition}

Direct successor constraints can also be formulated as atomic precedence constraints by adding the atomic constraint $c_i \triangleleft c_j$ to $\cal{A}$ and additional disjunctive constraints of the syntactic form ${\cal D}_2$ to specify that all other cavities must either come before or after the job pair $\langle c_i, c_j\rangle$, however, this formulation is less compact:
\begin{equation}
p(c_i) = p(c_j) +1 \leftrightarrow c_i \triangleleft c_j \wedge \forall \; c_l (i \neq l \neq j) \; c_l \triangleleft c_i \vee c_j \triangleleft c_l
\end{equation}

We now define a valid solution of a CTW instance as a solution, which satisfies all constraints defined for the instance.

\begin{definition}[Valid Solution]
Let $I$ be a CTW instance with $b$ two-sided and $n$ one-sided cables, \ie  $k=2b+n$ jobs, and sets of constraints $\cal{A}$, ${\cal A}_s$, $\cal{D}$, and $\cal{S}$. A permutation $\Per$ is a valid solution for $I$ if and only if  $\Per$ satisfies all constraints in ${\cal A} \cup {\cal D}  \cup {\cal S}$. Note that if $k=0$, all constraint sets are empty and any permutation of length 0 is a solution for the instance.
\end{definition}

As an illustrating example consider a CTW instance with cables $A$, $B$ and $C$, where $A$ and $B$ are two-sided cables defining the job pairs $\langle c_1, c_3\rangle$ and $\langle c_2, c_4\rangle$ respectively and $C$ is a one-sided cable defining the job $c_5$. This instance has parameters $k=5$ and $b=2$ and is subject to the following  constraints:
\begin{align*}
c_3 & \triangleleft c_4 \\
c_4 & \triangleleft c_1 \\ 
c_5 & \triangleleft c_4 \\
c_2 \triangleleft c_5 & \vee c_2 \triangleleft c_1\\
c_4 & \blacktriangleleft c_2 \\
\end{align*}
Of the $5!=120$ possible job permutations, only 8 are valid solutions satisfying all the constraints. One example of such a valid solution is the permutation $(c_5, c_3, c_4, c_2, c_1)$.

\subsection{Optimal Solutions of CTW Instances}\label{section:optimalSol}

In practice, one is not only interested in valid solutions, but in solutions of minimal cost. Four different cost functions are of interest. The first three cost functions aim to increase the robustness of the cable wiring actions by introducing penalties when a job pair is interrupted. This is since interrupting the processing of a job pair and storing cables into storage can increase the risk of the robot arm in getting caught in stored cables. The fourth function aims at keeping the violation of soft atomic constraints at a minimum, because a violation on the one hand allows the machine to make more flexible moves during wiring, but on the other hand it can impact robustness negatively. 

\begin{enumerate}
	\item $S=$ Number of interrupted job pairs, \ie cable ends that are temporally added to the cable storage by the machine to pick up another cable for insertion.
	\item $M=$ Maximum number of cables that are contained in storage simultaneously. 
	\item  $L=$ Longest time a cable end resides in storage expressed in terms of number of jobs.
	\item  $N=$ Number of violated soft atomic precedence constraints in ${\cal A}_s$.
\end{enumerate}

The formalization of the four criteria $S$, $M$, $L$ and $N$, makes repeated use of the indicator function $\mathbb{I}$. An indicator function $\mathbb{I}_{A}(x)$ for a set $A$ returns 1 when $x$ lies in the set $A$ and 0 otherwise. More formally, the indicator function $\mathbb{I}_{A}$ of a set $A$ is defined as
\begin{equation}
\mathbb{I}_A(x) =
\begin{cases}
1 & \text{if } x \in A\\
0 & \text{otherwise.}
\end{cases}
\end{equation}

Criterion $S$ counts how many of the job pairs were interrupted in a solution.

\begin{equation}
\label{s-cost}
S = 
\begin{cases}
0 & \text{if }  b = 0\\ 
\sum_{i= 1} ^{b} \mathbb{I}_{\bm{\mathsf{S}}}(i)        & \text{otherwise}
\end{cases}
\end{equation}
where $\bm{\mathsf{S}}=\{k \,\mid\,\, |p(c_k)-p(c_{k+b})| >1 \}$ and $\mathbb{I}_{\bm{\mathsf{S}}}(x)$ is the indicator function of the set $\bm{\mathsf{S}}$.

\medskip

To determine the criterion $M$, \ie the maximum number of cables that are stored simultaneously, we have to count for each job $c_l$ in the solution, how many job pairs exist where one end is plugged before $c_l$, but the other end is plugged after $c_l$, \ie how many job pairs are interrupted by a given job $c_l$.
\begin{equation}
M = 
\begin{cases}
0 & \text{if }  b = 0\\ 
\underset{\,l \in \{1, \dots, 2b+n\}} {\max} \sum_{i=1}^{b}   \mathbb{I}_{\bm{\mathsf{M}}_i}(l)   & \text{otherwise}
\end{cases}
\end{equation}
where $\bm{\mathsf{M}}_{j}=\{l \,\,\mid\,\, \min\{p(c_j), p(c_{j+b})\} < p(l) < \max\{p(c_j), p(c_{j+b})\} \}$ and $\mathbb{I}_{\bm{\mathsf{M}}_j}(l)$ is the indicator function of the set $\bm{\mathsf{M}}_j$. Note that in this case, the set $\bm{\mathsf{M}}_j$  depends on $j$.

\medskip

The optimization criterion $L$ is determined by measuring the number of jobs between job $c_i$ and job $c_{i+b}$ for each $i\in \{1, \dots, b\}$, \ie for each job pair, and taking the maximum value obtained for all job pairs. Note that if there are no jobs between two cable ends, the cable is plugged directly without accessing the storage.
\begin{equation}
L = 
\begin{cases}
0 & \text{if }  b = 0\\ 
\underset{\,i \in \{1, \dots, b\}} {\max} \; \left(|p(c_i) - p(c_{i+b}) | -1\right) & \text{otherwise}
\end{cases}
\end{equation}

\medskip

For criterion $N$, we count the number of violated soft atomic constraints from the set ${\cal A}_s$, so 
\begin{equation}
N =    \sum_{k \triangleleft \, l   \in \mathcal{A}_s}  \mathbb{I}_{\bm{\mathsf{N}}}(k,l)
\end{equation}
where $\bm{\mathsf{N}} = \{(x, y) \,\,\mid\,\, p(c_x) > p(c_y)\}$ and $\mathbb{I}_{\bm{\mathsf{N}}}$ is the indicator function of the set $\bm{\mathsf{N}}$.

\medskip

Assuming $k \neq 0$, all four criteria are bounded by a lower value of 0 and an upper value depending on the size of the solution $k= 2b+n$ as follows:

\begin{itemize}
	\item $0 \leqslant S \leqslant b < k$
	\item $0 \leqslant M \leqslant b < k$
	\item $0 \leqslant L \leqslant k-1 < k$
	\item $0 \leqslant N \leqslant \frac{k \cdot (k-1)}{2} < k^2$ as there can be  at most $\frac{k \cdot (k-1)}{2}$ soft atomic precedence constraints in a solution, otherwise the constraint graph contains a cycle  and the instance is unsatisfiable. 
\end{itemize}

As our objective function, we define the weighted sum of the four criteria $S$, $M$, $L$ and $N$. By the weight assigned to each criterion, we want to ensure that their possible values fall into non-overlapping intervals. This is to ensure that the solvers prioritize optimizing for $S$ before optimizing for $M$ and similarly prioritize optimizing for $M$ over optimizing for $L$ and so on. The value of $N$ for the instances in the benchmark set that we consider is more tightly bound than the worst possible bound, \ie in the benchmark set we always have $N < k$ instead of $N < k^2$. It is thus sufficient to weight the four objectives by powers of $k$ to eliminate the influence of one to the others. For all experiments in this paper, we therefore use the weighted sum\footnote{Alternativly, one can use lexicographic optimization~\cite{arora}, but not all solvers considered in this paper support it.} as defined in equation~(\ref{optkrit}) below as optimization objective:
 \begin{equation}\label{optkrit}
 k^3 \cdot S + k^2 \cdot M + k \cdot L + N 
 \end{equation}

For the example considered above and the solution $(c_5, c_3, c_4, c_2, c_1)$, this formula returns costs 161 $(S=M=N=1,  L=2)$, which makes this solution one of the two optimal solutions for this instance.

\subsection{NP-Hardness of the CTW Problem}

We now prove that the CTW problem is NP-hard by a reduction from the Maximum Acyclic Subgraph problem. Note that for this reduction, we assume the presence of soft atomic constraints and one-sided cables. 

\begin{theorem}[NP-hardness of CTW]\label{nphard}
The Cable Tree Wiring Problem (CTW) is NP-hard.
\end{theorem}

\begin{proofsk}
We prove NP-hardness by a reduction from the Maximum Acyclic Subgraph problem (MAS). 
Let us assume an MAS instance where we are given a directed graph $G = (V, E)$, where $V=\{1, \dots, n\}$ is the set of vertices of $G$ and $E$ is the set of directed edges of $G$. A solution to the MAS problem is a maximal set $E'\subseteq E$ of edges such that the resulting graph $G'=(V, E')$ is a directed acyclic graph. This problem is known to be NP-hard~\cite{karp}.

We now construct a CTW instance as follows. Let the set of vertices $V$ correspond to jobs of one-sided cables $c_1, \dots, c_n$. For each of the edges $e = (v, w) \in E$, where $v$ corresponds to job $c_{j}$ and $w$ to job $c_k$, we introduce a soft precedence constraint $c_j \triangleleft c_k$. The graph $G$ hence corresponds to the constraint graph of the CTW instance.

The solution to the CTW problem is then a permutation $\Per$ of the jobs with minimal cost, so with the fewest violated soft precedence constraints (since we have no pairs of jobs in this instance). Let $E'$ be the set of all edges corresponding to soft atomic precedence constraints that are not violated by the permutation. By the optimality of the solution for the CTW instance, the set $E'$ is the largest such set permitting a valid wiring sequence. Note further that a valid solution is possible if and only if the constraint graph has no cycles.

The set $E\setminus E'$ is hence the smallest set of edges that needs to be removed from $G$ to obtain a directed acyclic graph, and so $G'=(V,E')$ is the maximum acyclic subgraph of $G$. The solution to the CTW hence gives a solution to the Maximum Acyclic Subgraph problem. Since the solution size is clearly polynomial in the size of the MAS instance, it is a polynomial many-one reduction, proving that CTW is NP-hard.
\end{proofsk}

Since our proof relies on the presence of soft atomic constraints, the complexity of the CTW problem without soft atomic precedence constraints is still open. Note further, that NP-completeness of the corresponding CTW decision problem follows from Theorem~\ref{nphard}. For the special case that only (hard) atomic precedence constraints and no two-sided cables occur in an instance, the CTW problem can be solved in polynomial time. 

\begin{lemma}[CTW with only hard atomic constraints] A CTW instance with only one-sided cables and ${\cal A}_s = {\cal D} = {\cal S} = \varnothing$ can be solved in polynomial time.
\end{lemma}

\begin{proofsk}
First construct the constraint graph of the CTW instance by creating a vertex $i$ for every job $c_i$ and adding a directed edge from $i$ to $j$ to the graph if and only if the precedence constraint $c_i \triangleleft c_j$ exists in ${\cal A}$. By applying Kahn's algorithm~\cite{kahn1962topological} to the constraint graph, we obtain a topological ordering which is necessarily a valid permutation satisfying all the precedence constraints. Note that Kahn's algorithm has time complexity $O(k+e)$, where $k$ is the number of jobs and $e$ is the number of constraints in the CTW instance. The CTW instance can thus be solved in polynomial time.
\end{proofsk}

Based on the relationship to the Traveling Salesperson problem that we discuss in the next subsection, we conjecture that the CTW problem is NP-hard once two-sided cables are added even if only the set ${\cal A}$ of atomic precedence constraints is non-empty and all other constraint sets are empty. We also conjecture that the CTW problem restricted to containing only disjunctive precedence constraints is NP-hard. Disjunctive precedence constraints can be compiled away by converting the constraint set of disjunctive precedence constraints into disjunctive normal form (DNF) where each disjunct only contains atomic constraints. The compilation can lead to an exponential "blow-up" in the number of disjuncts only containing atomic precedence constraints~\cite{miltersen}.  Solution length remains polynomially bounded in this potentially larger search space, \ie membership in NP remains unchanged. Proving these conjectures is not straightforward as the CTW problem is a TSP variant with only tour-dependent edge costs, but no static edge costs, for which we could not find any formal complexity proofs.

\medskip

If we, however, restrict a CTW instance to only contain direct successor constraints, it can be solved in linear time.

\begin{lemma}[CTW with only direct successor constraints]
A CTW instance with $k=2b+n$ jobs and $b$ two-sided and $n$ one-sided cables where ${\cal A} = {\cal A}_s = {\cal D} = \emptyset$ and $ {\cal S} \not =\emptyset $ can be solved in linear time $O(k)$.
\end{lemma}

\begin{proofsk}
As a direct successor constraint can only be defined for an end of a two-sided cable, an optimal solution permutation with cost 0 can be constructed by first wiring all job pairs of the two-sided cables without interruption and then adding the jobs for the one-sided cables:
\[
\Per =  c_{1}, c_{1+b}, c_2, c_{2+b}, \dots, c_{b}, c_{2b}, c_{2b+1}, \dots, c_{2b+n}
\]
This sequence ensures that if a direct successor constraint is defined for some job $c_j$, either the other cable end is directly following or preceding it.

\end{proofsk}

\subsection{Relationship of CTW to TSP}
The CTW problem can be considered a variant of the traveling salesman problem (TSP) where cavities represent cities that need to be visited. TSP variants with precedence constraints have been studied in a number of papers with~\cite{kubo} being one of the first references, but see also~\cite{moon,rashid} for more recent overviews. Note that the CTW problem can also be considered as a variant of the pickup and delivery TSP problem, where the pickup and delivery locations can be exchanged with each other, but the tour should visit them directly. Closely related variations of the pickup and deliver problem have been studied~\cite{pickupTSP, routingTSP, pickupDelTSP}, but none of them is identical to the CTW variant. In the underlying graph, a city (job) is connected with any other city unless a precedence constraint $c_i < c_j$ is given, which removes the edge $c_j \mapsto c_i$ from the graph and also excludes all (sub)paths $c_j \dots \mapsto \dots c_i$ from the tour. Direct successor constraints remove even more edges from the graph, enforcing the edge $c_i \mapsto c_j$ as the only outgoing edge of $c_i$ in the graph. In the resulting partially connected graph, we need to compute a hamiltonian path, but not a hamiltonian cycle, because we do not need to return to the starting job after having visited each job exactly once. Computing a hamiltonian path on general graphs is NP-complete, however, on complete graphs, \ie in the unlikely case that no precedence constraints are given, the problem can be solved in linear time~\cite{rytter}.

Cost functions $S$, $M$, and $L$ aim at increasing robustness of the cable wiring actions, because interrupting the processing of a job pair and storing cables into storage can increase the risk of the robot arm in getting caught in stored cables. Note that the time it takes to complete each insertion is independent of its position in the schedule, because the robot arm of the machine needs to travel from the storage location where the cables are prepared to the cavity on the palette and back to storage after each insertion operation. Thus, travel times are constant and do not need to be considered in the problem formalization. This makes our problem at first glance different from routing problems, which usually minimize travel time. However, we will show now how to encode our storage costs as edge costs.

Recall, that we have defined the underlying graph of this problem as follows: Each node represents a job and two nodes are connected unless a precedence constraint $c_i < c_j$ is given, which removes the edge $c_j \mapsto c_i$. A direct successor constraint between $c_i$ and $c_j$ further removes all outgoing edges from $c_i$ except the edge $c_i \mapsto c_j$.
Given a job pair of a two-sided cable $\langle c_i, c_j \rangle$ and a further job $c_l$ where $l \neq i, j$, we define the edge costs as follows:
\begin{itemize}
\item Edges $c_i \mapsto c_j$ and $c_j \mapsto c_i$ have costs $0$, because we plug the two ends of the two-sided cable immediately after one another.
\item For an edge $c_i \mapsto c_l$ with $l \neq j$, costs are either $0$ or $1$ depending on whether the other end of the cable was visited on the tour to $c_i$ or not.
\begin{itemize}
\item If $c_j$ was visited on the tour to $c_i$, then edge $c_i \mapsto c_l$ has cost $0$ because $c_i$ is the ``second'' end of the cable that we take out of storage for plugging.
\item If $c_j$ was not visited, then edge $c_i \mapsto c_l$ has cost $1$ because we put $c_j$ into storage to work on the cable for $c_l$.
\end{itemize}
In the same way, costs are assigned to $c_j \mapsto c_l$. Intuitively, the edge cost is equal to $1$ if a  ``remaining" cable end is put into storage and a new cable is picked for wiring, otherwise the edge cost is $0$.
\end{itemize}

Let $1, \dots, k$ be the cable ends in a CTW instance with corresponding jobs $c_1, \dots, c_k$. Suppose further, as before, that $b$ of the cables are two-sided. Let $q(x)$ denote the inverse of the function $p$ introduced in Definition~\ref{solution}, \ie for a given position $x$, the function $q$ returns the job $c_j$ at position $x$ in the permutation sequence  ${\cal P}$.
We define a tour ${\cal P}$ in the graph to be a sequence of jobs (which correspond to nodes) $q(1), \dots, q(k)$, where for each $x \in \{1, \dots, k-1\}$ an edge between node $q(x)$ and $q(x+1)$ exists. A solution tour is Hamiltonian and visits every node in the graph exactly once. 

Let $\mathfrak{c}\left(q(x) \mapsto q(x+1)\right)$ be the cost of an edge $q(x) \mapsto q(x+1)$. $S$ is the total number of interrupted job pairs, so the number of edges with cost 1 occurring on the path. Hence, $S$ can be defined as

\begin{equation}
\label{s-cost-new}
S = \sum_{x=1}^{2b+n-1}\mathfrak{c}\left(q(x) \mapsto q(x+1)\right)
\end{equation}

We show that this definition of $S$ in Equation~\ref{s-cost-new} is equivalent to our original definition of $S$ in Equation~\ref{s-cost}. Equation~\ref{s-cost-new} can be rewritten as
\[ 
\hspace*{3mm}= \sum_{x=1}^{2b+n-1} \mathbb{I}_{K}(x)
\]
where $K=\{x \mid \text{ other end of the cable end corresponding to job } q(x) \text{ is put into storage}\}$, so this means $K$ is the set of all positions $x$ in the tour such that the edge $q(x) \mapsto q(x+1)$ has cost 1. An edge $q(x) \mapsto q(x+1)$ has cost 1 if and only if the other end of the cable end plugged in job $q(x)$ has not been plugged before $q(x)$ and is not plugged right after $q(x)$. This is then equal to 

\[ 
\hspace*{3mm} = \sum_{j=1}^{b} \sum_{x=1}^{2b+n-1} \mathbb{I}_{K^+_{ j}}(x) +  \mathbb{I}_{K^-_{j}}(x)
\]
where $K^+_{j} = \{x \mid q(x) = j \, \wedge \, x+1 < p(c_{j+b})\}$ and $K^-_{j} = \{x \mid q(x) =j+b \,\, \wedge \,\, x+1 < p(c_{j-b})\}$. So $K^+_j$ is the set containing the position in the scheduling round where the two-sided cable end $j$, where $j$ is at most $b$, is plugged and such that the other end of $j$ has not been plugged before $q(x)$ and is not plugged right after $q(x)$. The set $K^-_j$ is defined similarly for $j$ between $b+1$ and $2b$.
This can then be further simplified as follows

\[
\hspace*{3mm} = \sum_{j=1}^{b} \left(\sum_{x=1}^{2b+n-1} \mathbb{I}_{K^+_{j}}(x)\right) +  \left(\sum_{x=1}^{2b+n-1}\mathbb{I}_{K^-_{j}}(x)\right)
\]
which again can be simplified by joining the sets $K^+_{j}$ and $K^-_{j}$ for all $x$.

\[
\hspace*{3mm}= \sum_{j=1}^{b} \mathbb{I}_{S_{+}}(j) +  \mathbb{I}_{S_{-}}(j) 
\]
where $S_+ = \{j \mid p(c_j)+1 < p(c_{j+b})\}$ and $S_- = \{j \mid p(c_{j+b})+1 < p(c_j)\}$. Now since $S_+\cup S_- = S$, where $S=\{i \,\mid\,\, |p(c_i)-p(c_{i+b})| >1 \}$ is as defined in Section \ref{section:optimalSol}, we have that

\[
\hspace*{3mm}=\sum_{j=1}^{b} \mathbb{I}_{S}(j) 
\]
as required. This shows that the cable tree wiring problem can indeed be seen as a variant of the TSP problem combining precedence constraints and time-dependent edge costs.

\section{The \textsc{CTW} Benchmark Set}
\label{benchmark}

The \textsc{CTW} benchmark set comprises  205 real-world  and  73 artificial  instances of cable tree wiring problems. Each instance is defined by constants $k$ and $b$ and its constraint sets. The original cable tree, \ie the geometry of harnesses or cables is not contained in the instance description. Real-world examples originate from cable trees produced on the machines mostly for automotive applications. Artificial examples were constructed by industry specialists to highlight specific challenges when wiring cable trees during the development of the solution described in this paper. The benchmark set contains the following subsets:
\begin{itemize}
	\item satisfiable: a set of 71 artificial and 185 real-world instances,
	\item unsatisfiable: a set of 2 artificial and 20 real-world instances where the layout generates contradicting constraints, 
	\item challenge: a small subset of 5 artificial and 5 real-world  instances from the satisfiable set where solvers have difficulties finding an optimal solution. 
\end{itemize}

\begin{figure}[htb]
	\begin{center}	
		{\footnotesize
			\begin{tabular}{|l|r|r|r|r|r|r|r|}
				\hline 
				instance	& number of		&  atomic 		& soft atomic 	& disjunctive & sum of  & average  & maximum \\ 
				& two-sided cables $b$	&  constraints	& constraints 	& constraints &  constraints &  constrainedness & constrainedness\\ 	\hline 
				A033		&80				&418			&	74			& 224	& 756		&8.4			&34.0	\\ \hline
				A060		&100	 		&2471			&	86			& 339	& 2,946			&29.9			&89.5	\\ \hline
				A066		&170	 		&7651			&	156			& 842	& 8,734			&52.1			&150.0	\\ \hline
				A069		&186			&9313			&   172	     	& 971	& 10,549			&57.5			&166.0	\\ \hline
				A073		&198	 		&9870			&	184 		& 1211	& 11,364		&56.8			&168.0	\\ \hline
				R192		&104	 		&1270			&	74			& 197	& 1,593			&14.7			&64.5	\\ \hline
				R193		&104			&1056			&	86			& 99	& 1,293			&11.9			&43.5	\\ \hline
				R194		&112			&1471			&	81			& 204	& 1,812			&15.5			&68.5	\\ \hline
				R195		&110			&1525			&	77 			& 243	& 1,900			&16.7			&70.5	\\ \hline
				R196		&110			&1416			&	79			& 201	& 1,751			&15.3			&67.5	\\ \hline
			\end{tabular}
			\caption{\label{benchmark-overview}Parameters of the 10 challenge set instances.}
		}
	\end{center}	
\end{figure}

Figure~\ref{benchmark-overview} shows the number of two-sided cables and the number of atomic, soft atomic, and disjunctive constraints for each instance in the challenge set. No instance of the challenge set contains direct successor constraints or one-sided cables, but they occur in 80 of the other benchmark instances. To give an idea on how many constraints apply when choosing  a position value for a job in a permutation, we developed the notion of {\em constrainedness}, which is inspired by the clause/variable ratio used to  characterize the hardness of SAT instances~\cite{mitchell}. For each job, we determined the number of atomic or disjunctive constraints where the job occurs on the left-hand side of the  constraint. Occurrence in an atomic constraint counts as 1, occurrence in one disjunct within a disjunctive constraint counts as 0.5. We calculated the maximum, and average constrainedness over all jobs of an instance. The parameter gives a rough indication of the difficulty of an instance when considered together with the permutation length, \ie parameter $2b+n$. 

A better way of predicting the difficulty is to just determine the {\em sum of constraints} occurring in an instance. Note that we define the sum of constraints here as the sum of the number of two-sided cables, the number of atomic, soft atomic constraints, disjunctive constraints, and direct successor constraints. The number of two-sided cables is added to this sum as two-sided cables can also be viewed as soft direct successor constraints, because a penalty occurs if two ends of a two-sided cable are not plugged directly after one another. Using the sum of constraints, a correlation with the solving state for each solver can be observed, see Section~\ref{findings}. The sum of constraints in the CTW Benchmark set ranges from 0 to 11,766.  It is an interesting question of how these preliminary measures can be improved to accurately predict the difficulty of a CTW instance.  Note that the numbering of instances in the benchmark set does not reflect their difficulty in terms of the parameters discussed above.

The CTW benchmark set contains a variety of different instances. 20 instances require to compute permutations of length smaller than 10, 239 instances have a permutation length between 10 and 100, and 19 instances have a length of over 100 and up to 198 jobs. Real-world instances mostly range between 20 and 50 jobs with an average permutation length of 43. 40 instances contain one-sided cables ($k>2b$) and all of them except one (A008) are real-world instances. The three largest satisfiable instances A071, A072, and A073 contain around 10,000 atomic, over 1,000 disjunctive, and nearly 200 soft atomic constraints, but no direct successor constraints.  One of them, A073 is also part of the challenge set. 

The 22 unsatisfiable instances contain 2 artificial and 20 real-world instances. The largest of them have a permutation length between 70 and 80, over 1,000 atomic and around 150 disjunctive constraints. Average permutation length is 51 and the smallest unsatisfiable instance has permutation length 6. All of them contain direct successor constraints. Besides unsatisfiable instances, the benchmark set also contains a few pathological cases, such as for example instance R001 with no cables ($k=b=0$) and instance R002 with just a single one-sided cable ($k=1,b=0$).

The complete benchmark set with all models and instance data can be downloaded from \url{https://github.com/kw90/ctw_translation_toolchain}. Furthermore, this site also contains the code and documentation of the tool chain that we developed for the conversion of models and data into the different formats required by the various solvers, see also Section~\ref{models}. We also included an Excel file, which summarizes the parameters of all instances as well as all solver results that form the basis for  Section~\ref{solvers}. The CTW benchmark set has also been accepted for inclusion in the upcoming MiniZinc challenge.

\section{Modeling the CTW Problem}
\label{models}

All models are original work by the authors and are based on a so-called quadratic permutation representation of the TSP as described in~\cite{gutin}. The models are closely following the formalization of the problem to provide a base line for the empirical comparison of various solvers. This also implies that we represent disjunctive precedence constraints and direct successor constraints directly without rewriting them as described earlier. Other modeling variants can be imagined, \eg exploiting the relation to the TSP problem even more directly, modeling the problem directly as a scheduling problem, using a linear permutation representation, or a convex-hull encoding just to name a few. Our goal is a "natural" and straight-forward modeling of the problem to establish the desired base line rather than on achieving the best possible modeling, which would allow a solver to perform best on the problem.

In this section, we describe the constraint model $\mathbf{M}_\mathbf{C}$ in detail, which is the "native" model developed during the software project for the cable tree wiring solution. This model is used with the IBM Cplex CP constraint solver and written in OPL, the propriatary language of Cplex. Using OPL had the advantage that the model could be easily reviewed and discussed with domain experts, because of the compact and natural representation of data structures and constraints provided by OPL. Based on this model, a number of further modeling variants in other languages was developed in order to compare this model and the results obtained with Cplex CP with those from other solvers. In this section, we summarize the derivation process of all models and give a short overview on the main characteristics of the models. Their detailed description can be found in the appendix.

\subsection{The constraint model $\mathbf{M}_\mathbf{C}$}

The model $\mathbf{M}_\mathbf{C}$ uses domain variables to represent cavities and cables. We assign the cable end numbers to the cavities as there is exactly one cable plugged into one cavity and rather speak of cavities than cable ends in the model. For $k=2b+n$ cable ends/cavities with $b$ job pairs $\langle c_i, c_j\rangle$ numbered with $i = 1 \dots b$ and $j = i+b$ we introduce the corresponding ranges in the model. 
 
 {\scriptsize
 	\begin{verbatim}
 	int k  = ...;  //number of cavities, permutation length
 	int b = ...;   //number of two-sided cables, job pairs
 	range Cavities = 1..k;
 	range Cablestarts = 1..b;
 	\end{verbatim}
 }

The three sets of atomic, soft atomic and disjunctive constraints are explicitly introduced into the model. Cables that are too short for storage and that are subject to direct successor constraints are represented in a list of integers of cable end numbers.  
 
 {\scriptsize
 	\begin{verbatim}
 	{Atomic} AtomicConstraints = ...;
 	{Atomic} SoftAtomicConstraints = ...;
 	{Disjun} DisjunctiveConstraints = ...;
 	{int} DirectSuccessors = ...;
 	\end{verbatim}
 }

 Atomic and disjunctive constraints are represented as tuples
 
 {\scriptsize
 	\begin{verbatim}
 	tuple Atomic {int cbefore; int cafter;}
 	tuple Disjun {int c1before; int c1after;
 	int c2before; int c2after}
 	\end{verbatim}
 } 
 
The constraint model $M_C$ follows the approach by~\cite{smith} using two permutation sequences {\em cavity\_for\_position (cfp)} and {\em position\_for\_cavity (pfc)}. The {\em cfp} permutation assigns cavity identifiers to positions $1,\dots, 2b+n$ in the permutation, whereas the {\em pfc} permutation uses the cavity number as index and stores the position as value. Using this dual model for the permutation sequence was key in scaling the Cplex CP solver to larger CTW instances, which it could not solve using only one of the permutation representations. 

{\scriptsize
\begin{verbatim}
range Positions = 1..k;
range CavityPairs = 1..2*b ;
dvar int cfp[Positions] in Cavities;
dvar int pfc[Cavities] in Positions;

 \end{verbatim}
 }

For example, two permutations {\em cfp= 3,1,2} and {\em pfc=2,3,1} describe the same solution where cavity~3 is wired first, cavity~1 second, and cavity~2 last. The dual models are linked via a  channeling constraint $\forall j \in \mbox{\em Cavities}, p \in \mbox{\em Positions}: \; \mbox{\em pfc[j]} = p \Leftrightarrow  \mbox{\em cfp[p]} = j$ using the built-in {\em inverse} constraint in OPL. Furthermore, an {\em allDifferent} constraint is added for both permutation sequences to implement the constraints implied by Definition~\ref{solution}. Experiments on the influence of the two {\em allDifferent} constraints and the channeling constraint gave no clear picture. Different constraint combinations would increase or reduce solution time and costs on different instances. In the end, we decided to use the dual model with the three global constraints as our model.

{\scriptsize
	\begin{verbatim}
	allDifferent(pfc);  allDifferent(cfp); inverse(cfp, pfc);
	\end{verbatim}
}

All hard and soft constraints as well as optimization criteria are formulated on the {\em pfc} permutation following the experimental results in~\cite{smith}. Modeling the precedence and direct successor constraints is straightforward:

{\scriptsize
\begin{verbatim}
forall(c in AtomicConstraints) 
   pfc[c.cbefore] < pfc[c.cafter]; 	
	
forall(c in DisjunctiveConstraints) {
   pfc[c.c1before] < pfc[c.c1after] || 
     pfc[c.c2before] < pfc[c.c2after]; 
   if(c.c1before == c.c2before) 
   {maxl(pfc[c.c1after], pfc[c.c2after]) > pfc[c.c1before];}}
		
forall(i in DirectSuccessors: i<=b) 
   (pfc[i] < pfc[i+b]) => (pfc[i+b] - pfc[i] == 1);
	
forall(i in DirectSuccessors: i>b) 
   (pfc[i] < pfc[i-b]) => (pfc[i-b] - pfc[i] == 1);
\end{verbatim}
}

The definitions of the optimization criteria $S$, $M$, $L$ and $N$ read as follows:

{\scriptsize
\begin{verbatim}
dexpr int S = (b == 0) ? 0 : 
	 (sum(i in CableStarts) (abs(pfc[i] - pfc[i+b]) > 1))
	
dexpr int M = (b == 0) ? 0 :
	 (max(i in CavityPairs) (sum(j in CavityPairs: j<=b) 
	 ((pfc[j] < pfc[i] && pfc[i] < pfc[j+b]) ? 1 : 0) 
	 + sum(j in CavityPairs: j>b) 
	 ((pfc[j] < pfc[i] && pfc[i] < pfc[j-b]) ? 1 : 0)));
	
dexpr float L = (b == 0) ? 0 :	 
	 max(i in CableStarts) abs(pfc[i] - pfc[i+b]) - 1;
	
dexpr int N = sum(i in SoftAtomicConstraints)
	 (pfc[i.cbefore] > pfc[i.cafter]);

\end{verbatim}
}
	  	
The objective function is stated as	  	

{\scriptsize
	\begin{verbatim}	  		  	
minimize S * pow(k, 3)	+ M * pow(k, 2) + L * pow(k, 1) +  N;
\end{verbatim}
}

Alternatively, Cplex offers the built-in \textit{staticLex} function that defines a multi-criteria policy ordering the different criteria. However, in our experiments we found that Cplex performs slightly better when not using this function.  

Benchmark instances are represented in the .DAT format used by Cplex. The .DAT files used in the experiments with Cplex CP and MIP are directly exported from cable tree data in the Zeta machines using an XML-based software interface and an exporter written in C$\sharp$. Each file provides specific values for the integer parameters $k$ (number of jobs, permutation length) and $b$ (number of job pairs). Constraints are represented as sets of tuples of integer values enumerating the cavities. For example, instance R024 with 6 two-sided and  14 one-sided cables reads as follows (most constraints replaced by $\dots$)

{\scriptsize
	\begin{verbatim}		
k = 26;
	
b = 6;
	
AtomicConstraints = {<1,3>, <2,3>, <3,18>, <6,18>, <15,25>, <17,21>, ...}; 
	
SoftAtomicConstraints = {<2,1>, <4,3>, <6,5>, <12,26>, ...}; 
	
DisjunctiveConstraints = {<8,15,8,16>, <16,12,6,16>, <9,17,9,18>,...}; 
	
DirectSuccessors = {1,2,8,7,}; 
	
\end{verbatim}
}
Note that direct successor constraints are represented in a list of integers of cable end numbers, because they are specific to two-sided cables and express that once a cable end is plugged into a cavity, the other end must be plugged immediately after or must have been plugged before. The label of the other cable end is obvious from our numbering scheme using the $b$ parameter. Note that the integer $i$ occurring in the DirectSuccessors list means that the direct successor constraint $c_i \blacktriangleleft c_j$, where $j$ is the other end of the two-sided cable (so $j = i+b$ or $j=i-b$), exists in the problem instance. In the example above, we can see that both ends of the cables in job pairs $\langle c_1,c_7 \rangle$ and $\langle c_2,c_8 \rangle$ are too short for storage.

\subsection{Overview on Model and Data Variants and the Supporting Tool Chain}

\begin{figure}[htb]
	\begin{center} 
		\includegraphics[width=\columnwidth]{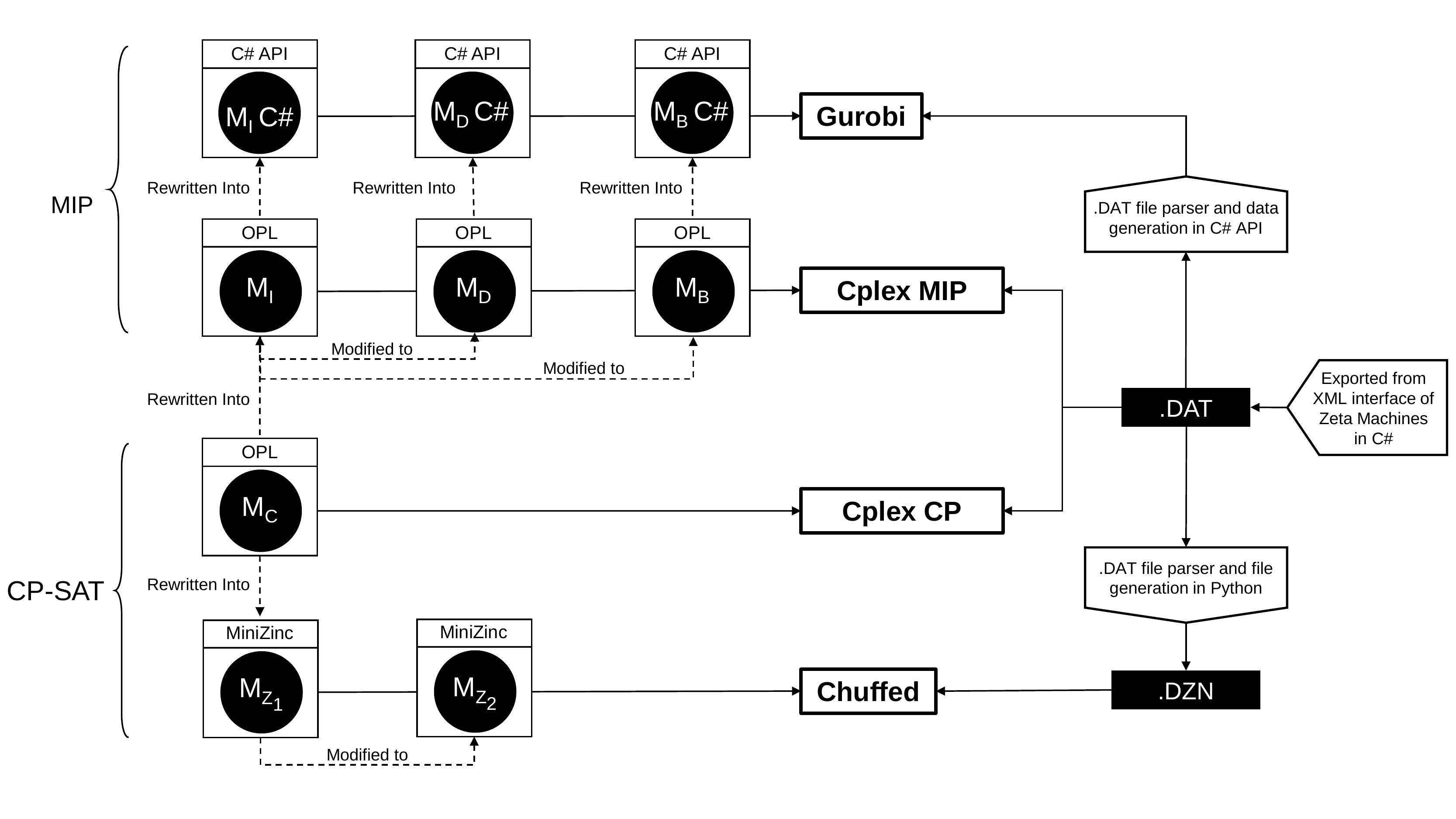}
		\caption{\label{fig:model-overview3a}Models and data formats for the Gurobi, Cplex CP, Cplex MIP, and Chuffed solvers that keep model and data in separate representations.} 
	\end{center}  
\end{figure}

Based on the model $\mathbf{M}_\mathbf{C}$, we derived several modeling variants and model/data representations to be able to perform a benchmarking using different solvers. As solvers use different modeling languages and data input formats, this turned out to be a very time-consuming manual process, which also included the necessity to write software to achieve the desired conversions. Figure~\ref{fig:model-overview3a} summarizes the derivation process for all solvers that separate model and instance data. It shows which model and data format is fed into which solver. In the following, we summarize the main characteristics of each model, a detailed description can be found in the appendix.

The constraint model $\mathbf{M}_\mathbf{Z_1}$ is an implementation of the $\mathbf{M}_\mathbf{C}$ model in the language MiniZinc~\cite{minizinc} used in experiments with the Chuffed and Google OR-Tools CP-SAT solvers. In contrast to the Cplex OPL language, the MiniZinc language does not provide the possibility to represent tuples. Therefore, OPL ranges are translated into sets of integers in MiniZinc and contraints are represented using arrays. Arguments of the arrays represent the cavities between which a constraint must hold. For disjunctive constraints, a 2-dimensional array is used. MiniZinc supports various global constraints, which are included into the model to express the {\em all\_different} constraint over the elements of an array in a straightforward way. The constraint model $\mathbf{M}_\mathbf{Z_1}$ does not make use of the dual representation of {\em pfc} and {\em cfp}, but only introduces the {\em position\_for\_cavity (pfc)} to represent the desired permutation sequence. In Section \ref{section:tuning}, we will however investigate the impact of adding this dual representation.

The model $\mathbf{M}_\mathbf{Z_2}$ is a variation of the $\mathbf{M}_\mathbf{Z_1}$ model and was created by Guido Tack, Monash University. In this model, disjunctive constraints are rewritten using an array of booleans and an additional constraint over this array is added. The array captures the truth value of the two disjuncts in a disjunctive constraint and the constraint states that at least one of the disjuncts must be true for the disjunctive constraint to be satisfied. Furthermore, an array of booleans is used to capture values of the optimization criteria, which are represented as sums over these boolean array values. 

For the Chuffed solver, the MiniZinc models $\mathbf{M}_\mathbf{Z_1}$ and $\mathbf{M}_\mathbf{Z_2}$ are working with benchmark instances represented in the .DZN format. These data files are generated using a Python script, which replace delimiters in the representation of the constraint sets as the .DAT and .DZN formats are quite similar.

Three variants of mixed-integer programming models $\mathbf{M}_\mathbf{I}$, $\mathbf{M}_\mathbf{D}$, and $\mathbf{M}_\mathbf{B}$ were manually written by us to run experiments with the Cplex MIP and Gurobi solvers. The models for Cplex MIP are represented in OPL and use the same .DAT files as the Cplex CP solver. In the MIP models, we rewrote constraints into equations and also replaced the {\em absolute} function in the criterion $L$ in order to express the criterion using only integers. The model $\mathbf{M}_\mathbf{I}$ follows the modeling approach from the $\mathbf{M}_\mathbf{C}$ model without any sophisticated rewriting of constraints as for example discussed in~\cite{refalo} in order to keep the model close to the constraint model and to avoid unintuitive reformulations. The model is based on the {\em pfc} permutation and represents constraints as tuples of integers in the same way as the $\mathbf{M}_\mathbf{C}$ model. It does not make use of the dual {\em cfp} representation, and thus needs no channeling constraint. The {\em allDifferent} constraint is replaced by pairwise inequalities over integer cavity numbers assigned to {\em pfc} positions. All other constraints rewrite the $<$ condition over permutation values from the $\mathbf{M}_\mathbf{C}$ model as inequalities. The model $\mathbf{M}_\mathbf{D}$ extends the model $\mathbf{M}_\mathbf{I}$ with the dual {\em cfp} representation, inequality constraints for {\em cfp} and two channeling constraints, which formulate pairwise equality conditions.

The model $\mathbf{M}_\mathbf{B}$ makes use of a big-M reformulation for the disjunctive constraints~\cite{ruiz} using additional decision variables and non-integer (floating point) optimization costs. An upper triangular matrix of booleans is used to represent that one cavity is plugged before another one. Constraints can be directly expressed in this matrix representation and the solution can be read off the matrix. Additional decision variables are used to capture the minimum and maximum positions of the cavities in the permutation. Their values are set by additional constraints.

The variants of the  $\mathbf{M}_\mathbf{I}$, $\mathbf{M}_\mathbf{D}$, and $\mathbf{M}_\mathbf{B}$ models for the Gurobi MIP solver are  implementations in the C$\sharp$ API provided by this solver. To feed the instance data into this solver, a .DAT file parser was implemented in C$\sharp$ such that the data can be directly generated in the C$\sharp$ API of Gurobi. In contrast to Cplex OPL, inequalities cannot be directly expressed in the Gurobi API, but must be rewritten as linear inequality expressions, which required us to introduce additional binary variables for each inequality constraint. Similarly, disjunctive constraints as well as the {\em allDifferent} constraint are rewritten using binary variables. In addition, we had to introduce additional variables for the optimization criteria, which capture if the end of a cable comes first or second in the permutation sequence.

\begin{figure}[htb]
	\begin{center} 
		\includegraphics[width=\columnwidth]{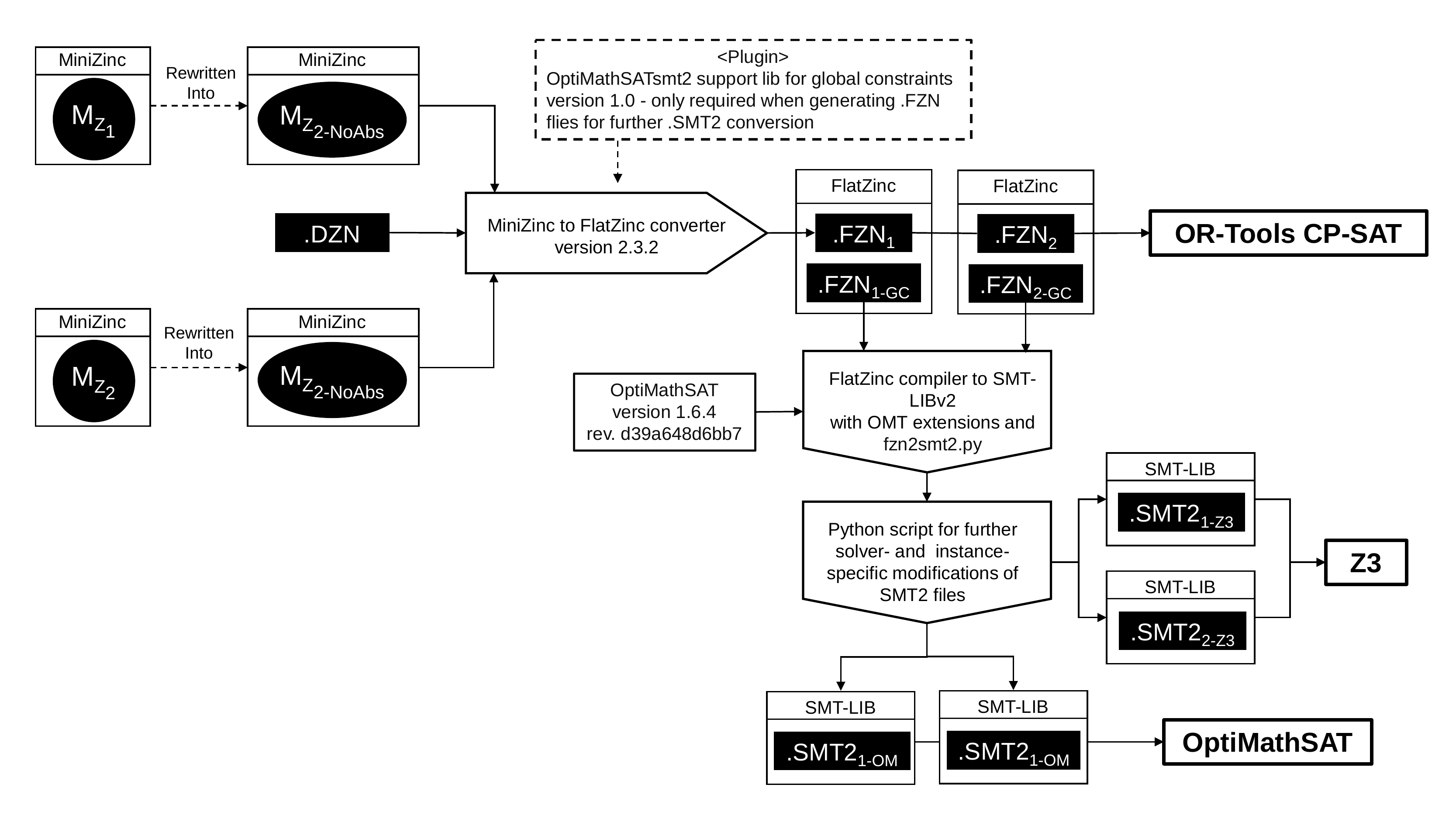}
		\caption{\label{fig:model-overview3b}Tool chain for the Google OR-Tools, Z3, and OptiMathSAT solvers that use a single integrated model-data file for each instance.} 
	\end{center}  
\end{figure}

Figure~\ref{fig:model-overview3b} summarizes the tool chain for the experiments with the Google OR-Tools CP-SAT solver and the two OMT solvers Z3 and OptiMathSAT, which use a single integrated model-data file for each instance.  First, the MiniZinc models $\mathbf{M}_\mathbf{Z_1}$ and $\mathbf{M}_\mathbf{Z_2}$ are rewritten into MiniZinc models $\mathbf{M}_\mathbf{Z_1-NoAbs}$ and $\mathbf{M}_\mathbf{Z_2-NoAbs}$, because the MiniZinc to Flatzinc converter does not support the {\em absolute} function. Second, using the data files in Chuffed's .DZN format and one of the rewritten MiniZinc models, we generate two sets of FlatZinc files .FZN$_1$ and .FZN$_2$ in .FZN format, which provide the input into the Google OR-Tools CP-SAT solver.

In order to generate the .SMT2 files for the OMT solvers, this tool chain needs to be invoked again, but this time additionally using the OptiMathSATsmt2 support library marked with dashed lines in Figure~\ref{fig:model-overview3b}. This library generates another variant of the models and instance data named .FZN$_{1-GC}$ and .FZN$_{1-GC}$ supporting global constraints. The .FZN$_{1-GC}$ and .FZN$_{1-GC}$ files can then be further processed  with a FlatZinc to SMT2 compiler using OptiMathSAT, specific syntax support for Z3 and OptiMathSAT, and OMT extensions~\cite{trentin,trentin2}. Some information from our models is, however, not translated correctly to these files requiring postprocessing of the files with our own Python script. First, the lower and upper bounds for the decision variable \textit{pfc} are missing. Our scripts therefore need to add functions to the .SMT2 file setting the bounds larger than zero and lower or equal to the length of the permutation $k$. For example, for a permutation of length 20, these functions read as follows:

{\scriptsize
	\begin{verbatim}
(define-fun lbound20 () Bool (> @pfc@20 0))
(define-fun ubound20 () Bool (<= @pfc@20 20))
	\end{verbatim}
}

The two functions have to be true in our model:

{\scriptsize
	\begin{verbatim}
	(assert lbound20)
	(assert ubound20)
	\end{verbatim}
}

Second, the extraction of the \textit{pfc} sequence from the solution model lacks a clear naming scheme. The sequence to be extracted corresponds to the first $k$ variables starting with the name {\texttt X\_INTRODUCED}. A workaround adds the relevant variables as comments to the output file. These comments are then used by our solution extractor to correctly extract the permutation values. For example if $k=4$, the  comments in the beginning of the file read:

{\scriptsize
	\begin{verbatim}
;; k=4
;; Extract pfc from
;; X_INTRODUCED_0_
;; X_INTRODUCED_1_
;; X_INTRODUCED_2_
;; X_INTRODUCED_3_
	\end{verbatim}
}

For each model, we obtained two sets .SMT2$_{Z3}$ and .SMT2$_{OM}$ of files in .SMT2 format, which are specific to Z3 and OptiMathSAT. A project is available on github to access this elaborate tool chain for the generation of the .SMT2 formats for the two OMT solvers, see \url{https://github.com/kw90/ctw_translation_toolchain}. The repository contains a Docker environment specified with scripts for installing all dependencies from the  OptiMathSAT, Z3, libminizinc, and fzn2omt sources. A Jupyter notebooks automatically proceeds over all files in the specified directory, applies the translation and the necessary adjustments. For the experiments with OptimathSAT, we forked a project on github \url{https://github.com/kw90/omt_python_timeout_wrapper}, which implements a Python wrapper to call the OptimMathSAT C library, such that we were able to run the solver with a given time limit and extract the best solution found. In contrast to the other solvers, which run directly under Windows 10, OptiMathSAT runs on a Linux Ubuntu 20.04 machine within the Windows Subsystem for Linux WSL.

\section{Benchmarking Models and Solvers on the CTW Problem}
\label{solvers}

In the following, we summarize our findings from experiments with the various solvers and models. All experiments were run on a Windows 10 virtual machine with four 2.30 GHz processors and 8 GB memory. All solvers are called from our own C$\sharp$ code environment, which performs all necessary data and model conversions and result validation. We tested all solvers in their default configuration settings, which often means that a solver automatically performs strategy selection. For Chuffed and Google ORT-Tools we used their freesearch option as default configuration.For Z3, a specific search strategy has to be selected and we used its default solver for weighted MaxSAT problems called MaxRes~\cite{maxres1,maxres2}. For OptiMathSat, we used its  OMT-based encoding and engine setting as default strategy. For Z3, we also experimented with other search strategies. For the constraint solvers, we investigated different settings of propagation levels for Cplex CP and ran some experiments using search annotations in MiniZinc for Chuffed and OR-Tools. All solvers were tested on the entire benchmark set in a single run, except Gurobi, which we tested in chunks of 60 instances due to out-of-memory problems that we could not resolve otherwise. 

The extraction of solution data is specific for each solver. We used their generated logs and/or API to access the solver state and solution. Each solution was validated for correctness using our own software, which checks that a generated permutation sequence does satisfy all constraints. The software also recalculates the values of the optimization criteria and the overall objective. When showing solution costs, we used these recalculated values to make the solution costs comparable across different solvers as single cases of deviations occurred for some instances. We discuss these in more detail when we summarize our findings at the end of this section. 

The numbers we report in the following are all from  a single run of a solver for consistency reasons. We performed up to three runs for each solver and noticed minimal differences in the number of solved instances, solution costs or runtime, however felt that selecting numbers from a single (in our case the first) run gives a better impression than computing the average from 3 runs. For example, some solvers can solve up to two additional instances, but given the total subset of 256 instances, this difference is very small and does not change the overall picture.

\subsection{Mapping Proprietary Solver States to a Uniform Set of Result States}

Before beginning any evaluation, we needed to decide how to map the individual solver states to a common set of states. For the subsequent experiments, we are interested in four possible outcomes of a solver on a benchmark instance: {\em unsatisfiable}, meaning that the solver has proven the instance to have no solution,  {\em optimal}, meaning that the solver has found a solution and proven that no better solution with lower costs exists, {\em suboptimal}, meaning that the solver has found a solution, but was not able to prove it as optimal, and finally, {\em unsolved}, meaning that the solver was not able to find a solution or prove an instance as being unsatisfiable within a given time limit. Any other state returned by a solver is mapped to {\em undefined} and we discuss these cases in more detail at the end of this section. 

Table~\ref{solverstates} summarizes the mapping. For Chuffed and OR-Tools, the entries refer to a string syntax used by MiniZinc to represent the status of a solution. For Cplex CP, solver states are defined by a parameter value IloCP::ParameterValues of the solver and there is a separate boolean parameter to indicate whether a solution was found or not. For Cplex MIP, we check the solver state in the Cplex.CplexStatus parameter and the existence of a solution. Gurobi returns a numerical solver status code in parameter GRB.Status and a solution count in parameter GRBModel.SolCount that we check in addition to the status code. Z3 distinguishes three different solver states in a status variable and adds the state in a string to the output file containing the solution. However, it has no explicit state for marking a solution as being optimal. Therefore, we checked its solution costs and if this is equal to the cost of a solution marked as optimal by another solver we also count it as an optimal solution found by Z3. OptiMathSAT outputs a solution status on the Linux console, to which we added a timeout output string via our Python wrapper in case the solver exhausted the time limit.

\begin{figure}[htb]
\begin{center}	
{\footnotesize
\begin{tabular}{|l|p{2cm}|p{2.5cm}|p{3.8cm}|p{3.3cm}|}  \hline   
	&  \multicolumn{1}{c|}{ {\bf optimal}} &  \multicolumn{1}{c|}{{\bf suboptimal}} &  \multicolumn{1}{c|}{{\bf unsatisfiable}} &  \multicolumn{1}{c|}{{\bf unsolved}}\\ \hline
Chuffed   & "==========" & "----------" & "=====UNSATISFIABLE=====" & "=====UNKNOWN=====" \\ \hline
OR-Tools  & "==========" & "----------" & "=====UNSATISFIABLE=====" & "TIMEOUT" \\ \hline
Cplex CP  &  SearchCompleted &  SearchStopped & SearchCompleted &  SearchStopped \\
& and & and & and & and \\
 & cp.Solve == true  & cp.Solve == true  & cp.Solve == false  &cp.Solve == false \\ \hline
Cplex MIP &	Optimal & AbortTimeLim & & AbortTimeLim \\
          &   or    & and & Infeasible  & and \\
          & OptimalTol & cplex.Solve == true &  & cplex.Solve == false \\ \hline
Gurobi    & GRB.Status==2   &GRB.Status==9 & GRB.Status==3	 & GRB.Status==9 \\
          &              & and        &              & and \\
          &              & SolCount $>$ 0 &   & SolCount == 0 \\ \hline         
Z3        &  n/a (cost-based)          & "sat" & "unsat"  & "timeout" \\ \hline	
OptiMathSat &  "sat\_optimal" & "sat" & "unsat" & "timeout"\\ \hline
\end{tabular}
		}
		\caption{\label{solverstates} Mapping of solver-specific information about solver state and solution existence to our four  outcomes of a benchmark test. Any other state returned by a solver is mapped to a state {\em undefined}.}
	\end{center}	
\end{figure}

\subsection{Finding an Appropriate Time Limit for the Experiments}

The second question is how long to invoke a solver on an instance. We are interested in setting a time limit, which allows solvers to find good solutions or even solve instances optimally. However, solvers can easily get stuck in large search spaces resulting from the very large instances in our benchmark set and investing more time will not allow them to significantly improve solution quality. Therefore, we ran a number of tests with the instances from the challenge set using time limits of 2, 5, 10, and 20 minutes. For these tests, we selected all constraint solvers, the Cplex MIP solver, and Z3. We compared the costs of solutions found by these solvers under different time limits. Table~\ref{fig:timelimit-cplex} summarizes the results for the Cplex CP solver using the $\mathbf{M}_\mathbf{C}$ model, which solves all challenge set instances under all time limits, but cannot prove any of its solutions as optimal.

\begin{figure}[htb]
\begin{center}	
{\footnotesize
\begin{tabular}{|l|r|r|r|r|}  \hline   
	& \multicolumn{4}{c|}{Cplex CP  $\mathbf{M}_\mathbf{C}$}      \\
	Instance     & 2  minutes & 5 minutes    & 10 minutes 	& 20 minutes    \\ \hline 
	A033 & 7,742,021  & 7,742,021	 & 7,741,861    & 7,741,781	    \\ \hline
	A060 & 43,318,041 & 43,308,025   & 42,338,828   & 41,317,903	\\ \hline
	A066 & 374,829,658 & 364,859,160 & 359,918,105  & 335,380,634	\\ \hline
	A069 & 529,386,562 & 503,541,657 & 484,268,711  & 464,931,037	\\ \hline
	A073 & 630,266,347 & 536,520,241 & 420,367,496  & 288,011,609	\\ \hline
	R192 & 26,007,513  & 22,587,155  & 20,337,530   & 20,337,530	\\ \hline
	R193 & 15,827,902 & 10,191,828   & 10,191,619   & 10,181,118	\\ \hline
	R194 & 36,724,491 & 28,217,186   & 19,748,757	& 19,748,757	\\ \hline
	R195 & 34,748,796 & 28,069,153	 & 28,069,153	& 28,069,150	\\ \hline
	R196 & 34,782,897 & 32,105,826 	 & 30,774,826	& 18,685,725	\\ \hline
\end{tabular}
}
\caption{\label{fig:timelimit-cplex} Results for Cplex CP on the challenge set under different time limits. Entries show solution costs. None of the solutions found was marked as optimal.}
\end{center}	
\end{figure}

\begin{figure}[htb]
	\begin{center} 
		\includegraphics[width=0.8\columnwidth]{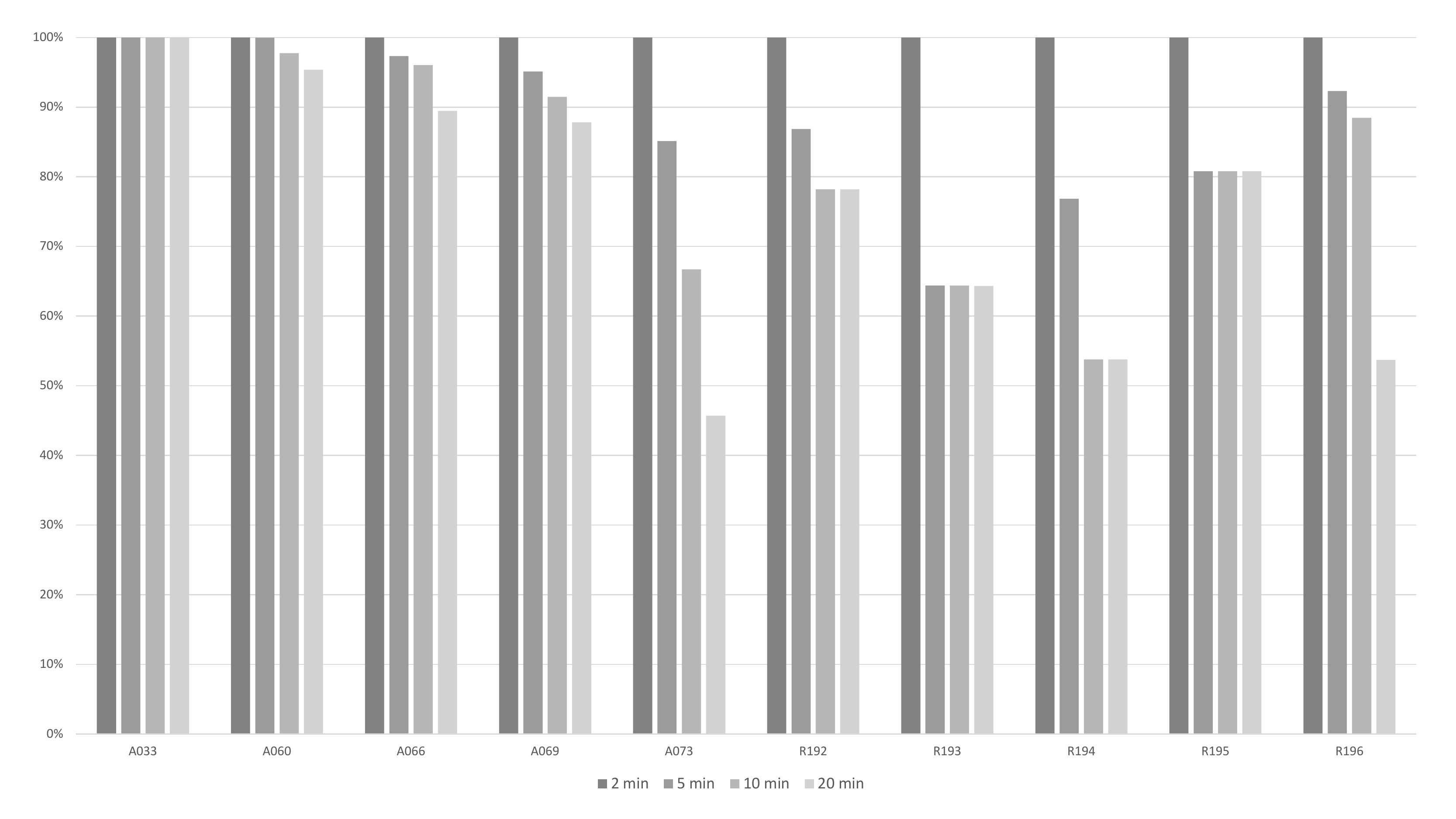}
		\caption{\label{fig:tl-cplex}Relative decrease in solution costs for Cplex CP on the challenge set over all tested time limits. Costs of the solution found within 2 minutes is set as 100 \%.} 
	\end{center}  
\end{figure}

Cplex CP can improve solution costs on all instances when given more time. However as Figure~\ref{fig:tl-cplex} illustrates, the improvement is much less significant from 10 to 20 minutes when compared with the decrease in costs made when going from 2 to 5. Investing 10 minutes yields relevant cost reductions for only three instances.  

Table~\ref{fig:timelimit-other} summarizes the results for all other solvers, which only solve a few instances from the challenge set. Only OR-Tools is able to solve one instance optimally in 5 minutes and 2 more instances optimally in 20 minutes. Cplex MIP was only able to solve one instance, whereas Z3 was not able to solve any instance even when given a 20 minutes time limit. 

\begin{figure}[htb]
\begin{center}	
{\footnotesize
\begin{tabular}{|l|r|r|r|r|r|}  \hline   
	Instance	 & 2  minutes  & 5 minutes  & 10 minutes	& 20 minutes & \multicolumn{1}{c|}{Solver}		 \\ \hline 
	\multirow{6}{*}{A033}	 
	&  & 20,671,476 & 19,634,521 &	18,610,991 & Chuffed $\mathbf{M}_\mathbf{Z_1}$  \\ 	
	& 19,660,117 &20,099,646 & 19,615,230 &18,563,257 & Chuffed $\mathbf{M}_\mathbf{Z_2}$ \\ 
	& 	      &     & 18,026,120   &	 16,465,075	&Cplex MIP  $\mathbf{M}_\mathbf{I}$	\\ 
	&  & \textbf{5,153,238} &  5,153,238 &  5,153,238 & OR-Tools FZN$_1$ \\
	& 11,391,543 & \textbf{5,153,238} & 5,153,238 & 5,153,238 & OR-Tools FZN$_2$ \\ \hline       
	A060 & 	      &     &  &	46,359,032	&Cplex MIP  $\mathbf{M}_\mathbf{I}$	\\ \hline
	\multirow{2}{*}{R192} 
	&  	& 		& 	58,965,217 &	58,965,217 & Chuffed $\mathbf{M}_\mathbf{Z_1}$	\\
	&       &      & 28,202,636  & \textbf{12,408,258} &  OR-Tools FZN$_1$ \\
	&       &      &  36,118,707   &  \textbf{12,408,258} &  OR-Tools FZN$_2$ \\ \hline
	\multirow{3}{*}{R193}  
	&			&  	& 		& 	54,215,758 & Chuffed $\mathbf{M}_\mathbf{Z_2}$	\\
	&			&  	& 		&  11,284,655  & OR-Tools FZN$_1$\\
	&			&  	& 		&   19,234,102 & OR-Tools FZN$_2$\\ \hline
	\multirow{2}{*}{R194} 
	&  	& 		& 		&	78,936,063 & OR-Tools FZN$_1$\\ 
	&  	& 		& 	49,328,866 & 25,369,932 & OR-Tools FZN$_2$\\\hline
	\multirow{3}{*}{R195} 
	&  	& 	& 	&	 73,481,904 & Chuffed $\mathbf{M}_\mathbf{Z_2}$\\
	&  	& 	& 	&    28,107,669 & OR-Tools FZN$_1$\\ 
	&  	& 	& 	&   26,750,039 & OR-Tools FZN$_2$\\ 				\hline
	\multirow{2}{*}{R196} 
	& 		& 		& 		&	24,021,937	 & OR-Tools FZN$_1$	\\ 
	& 		& 	61,487,932 & 	40,095,576 &	\textbf{14,679,739} & OR-Tools FZN$_2$ \\\hline
\end{tabular}}

		\caption{\label{fig:timelimit-other} Summary of results obtained by Chuffed, Cplex Mip and OR-Tools on the challenge set under different time limits. Only those instances are shown where at least one solution was found by a solver. Entries show solution costs. The few optimal solutions found are marked in bold. Cells with no entries mean that no solution was found by any of these solvers.}
	\end{center}	
\end{figure}


As our experiments show, a 5 minute time limit allows solvers to find solutions. For instances, for which only suboptimal solutions are found, running the solvers for 10 or 20 minutes yields improvements, but they rarely allow solvers to find optimal solutions. We thus set the time limit for all subsequent experiments to 5 minutes. This limit is also an acceptable limit from a practical and application-oriented perspective as computing the control for a machine does not need to happen instantly and 5 minute cycles are acceptable for users.

\subsection{Solver Performance on CTW Benchmark Set}


In the following, we discuss our findings on running all solvers in their default configuration on the entire benchmark set of 256 satisfiable and 22 unsatisfiable instances. 

\begin{figure}[htb]
	\begin{center}	
		{\footnotesize
			\begin{tabular}{|l|r|r|r|r|r|}  \hline 
				& Cplex CP & \multicolumn{2}{|c|}{Chuffed} & \multicolumn{2}{|c|}{OR-Tools}  \\ 
				Solver State & $\mathbf{M}_\mathbf{C}$ &  $\mathbf{M}_\mathbf{Z_1}$ &  $\mathbf{M}_\mathbf{Z_2}$ 
				& $\mathbf{FZN_1}$ & $\mathbf{FZN_2}$   \\  \hline	
				optimal  				& 139 (+54) 	& 144 	& 135 	& 224	&  228		\\     	
				suboptimal  			& 117 	& 85 	& 88	&	6	&    5		\\ 		 
				unsolved  				& 0 	& 26 	& 32 	&  24 	&   21		\\ 
				unsatisfiable   		& 22 	& 22	& 22	&  22	&  	22	\\  	
				undefined    			& 0 	& 1 	& 1		&   2	&	2	\\  \hline				
				\textbf{TOTAL Solved}	& 256	& 229	& 223	& 230	&  	233	\\ \hline	
			\end{tabular}
		}
		\caption{\label{result-cpsat}Performance of constraint solvers in default configuration on entire benchmark set.}
	\end{center}	
\end{figure}

Table~\ref{result-cpsat} summarizes the results for the constraint solvers using different model variants. Only Cplex CP finds solutions for all satisfiable instances, the other constraint solvers encounter between 21 and 32 unsolvable instances. Besides the 139 instances, for which Cplex can prove optimality of its solution, it finds 54 solutions of minimal cost, but cannot prove them as optimal. Other solvers proved their solutions with the same costs as optimal and therefore, we show this number in parentheses among the optimal solutions. On the 6 instances, which OR-Tools using the .FZN$_1$ model cannot solve optimally, no other solver finds an optimal solution. However, when using the .FZN$_2$ model  OR-Tools solves 3 of these 6 instances optimally. On the 5 instances, which OR-Tools solves suboptimally using the .FZN$_2$ model, Cplex CP finds suboptimal solutions of lower costs for all of them. On some instances, Chuffed and OR-Tools end up in an undefined state, which we discuss in more detail when we summarize our findings at the end of this section. All solvers identify the unsatisfiable 22 instances instantly.

\begin{figure}[htb]
\begin{center}	
{\footnotesize
\begin{tabular}{|l|r|r|r|r|r|}  \hline 
Benchmark  Subset & Cplex CP & \multicolumn{2}{|c|}{Chuffed} & \multicolumn{2}{|c|}{OR-Tools}  \\ 
cost \& runtimes & $\mathbf{M}_\mathbf{C}$ &  $\mathbf{M}_\mathbf{Z_1}$ &  $\mathbf{M}_\mathbf{Z_2}$ 
& $\mathbf{FZN_1}$ & $\mathbf{FZN_2}$   \\  \hline	
\textbf{Total Solved (218 instances)}& 	&  &  & 	&  	\\ 
total costs     & 165,861,766	& 591,651,075 & 581,427,792 & 182,055,350		&  162,485,868	\\     	
relative costs 	(in $\%$) &  102	& 364	& 358	& 112		& 	100	\\ 	\hline	 
\textbf{Solved Optimally (123 instances)} & 	&  &  & 	&  	\\ 
total costs 		& 3,930,031 	&   3,930,014	&   3,930,014	 &   3,930,014	 &   3,930,014	\\   
total runtimes (in s) 	&  1466 &  3295 &   3128 &  303 &  307  	\\  	
relative runtimes (in $\%$)	 & 483 	&   1087	&  1032	 &  100	 & 101 \\  \hline	
				
			\end{tabular}
		}
		\caption{\label{costs-cpsat}Costs and runtimes for optimal and suboptimal solutions returned by constraint solvers in default configuration.  Percentages are rounded mathematically to the next integer value with setting the best-performing solver to 100 \%.}
	\end{center}	
\end{figure}

There is a subset of 218 instances that all solvers using any of their models can solve optimally or suboptimally for which we compare solution costs and runtimes in Figure~\ref{costs-cpsat}. This subset of 218 instances has an average permutation length of 38 with the number of both-sided cables being 19 on average. These instances contain on average 229 atomic, 26 soft atomic, 43 disjunctive, and 3 direct successor constraints. The largest instance in this set, R189,  has permutation length 100, it comprises 44 both-sided cables and contains 1305 atomic, 62 soft atomic, and 133 disjunctive constraints. As Figure~\ref{costs-cpsat} summarizes, the cost of solutions returned by OR-Tools and Cplex CP is in a similar range, whereas the cost of solutions returned by Chuffed is significantly higher. 
Comparing runtimes for this subset is not interesting as solvers exploit the time limit of 5 minutes when finding a suboptimal solution. Among the 218 instances, all solvers can find optimal solutions for 123 of them. For this subset, the average permutation length is 22 and only very few instances contain one-sided cables. On average, these instances  contain 60 atomic, 14 soft atomic, 13 disjunctive, and 2 direct successor constraints. The largest instance in this subset, R197,  has permutation length 49, contains 24 both-sided cables, and comprises 343 atomic, 31 soft atomic, 31 disjunctive, and 3 direct successor constraints. As one would expect, all solvers return optimal solutions of identical costs except Cplex CP, for which slightly higher costs are shown. We discuss this rather surprising deviation at the end of this section. It is worth looking at the runtime solvers need to find optimal solutions in this subset using their respective (and different) models. OR-Tools using the $\mathbf{FZN_1}$ and $\mathbf{FZN_2}$ models is the fastest and needs only a little bit more than 5 minutes, followed by Cplex CP using the $\mathbf{M}_\mathbf{C}$ model with a little bit more than 24 minutes. Chuffed needs more than one hour for finding optimal solutions using the $\mathbf{M}_\mathbf{Z_1}$ and $\mathbf{M}_\mathbf{Z_2}$ models and is thus 10 times slower than OR-Tools. 


Table~\ref{result-mip} summarizes the results for Cplex MIp and Gurobi on the three different mixed-integer models.

\begin{figure}[htb]
\begin{center}	
{\footnotesize
\begin{tabular}{|l|r|r|r|r|r|r|}  \hline 
 &  \multicolumn{3}{c|}{Cplex MIP}    & \multicolumn{3}{c|}{Gurobi}  \\ 
Solver State & $\mathbf{M}_\mathbf{I}$ & $\mathbf{M}_\mathbf{D}$ & $\mathbf{M}_\mathbf{B}$ 
& $\mathbf{M}_\mathbf{I C\sharp}$ & $\mathbf{M}_\mathbf{D C\sharp}$ & $\mathbf{M}_\mathbf{B C\sharp}$  \\  \hline	
optimal  				& 126 	& 82	& 126 	& 134 & 103 & 134  \\     	
suboptimal      		& 77	& 56	& 24	& 17  & 25  & 19 \\ 		 
unsolved  				& 46	& 118	& 106	& 104 & 127 & 102\\ 
unsatisfiable   		& 21	& 21	& 19	& 23  & 23  & 23	\\ 	
undefined    			& 8		& 1		& 3 	& 0   & 0   & 0\\  \hline			
\textbf{TOTAL Solved}	& 203	& 138	& 150   & 151 & 128 & 153\\ \hline		
\end{tabular}
}
\caption{\label{result-mip}Performance of MIP solvers in default configuration  on entire benchmark set.}
	\end{center}	
\end{figure}

As the entries in Table~\ref{result-mip} shows, the two MIP solvers show similar behavior in the number of found optimal solutions, but the dual model  $\mathbf{M}_\mathbf{D}$ yields significantly fewer solved instances. Cplex MIP finds suboptimal solutions for many more instances than Gurobi.  There is a subset of 150 instances where both solvers find an optimal or a suboptimal solution using the $\mathbf{M}_\mathbf{I}$ model. For this subset, the total solution cost for Gurobi is 15,531,675, whereas it is 22,309,566, \ie 43\% higher for Cplex MIP. Using the 
$\mathbf{M}_\mathbf{B}$ model, both solvers find solutions for 141 instances, but for this set the total solution cost is 9,786,394 for Cplex MIP, whereas Gurobi returns total costs of 16,717,753, \ie  71\% higher for Gurobi. 

A significantly higher number of instances remains unsolved by the Cplex and Gurobi MIP solvers when compared to the constraint solvers. Gurobi identifies the 22 unsatisfiable instances, but also returns state \textit{unsatisfiable} for instance R001, which contains no cables. 
Depending on the model, Cplex MIP identifies 21 or 19 of the unsatisfiable 22 instances. With the $\mathbf{M}_\mathbf{I}$ model, it returns the solution state \textit{undbounded or infeasible} on 8 instances, which we map to our undefined state. One of these instances is from the set of 22 unsatisfiable instances, the remaining 7 are satisfiable.  With the $\mathbf{M}_\mathbf{D}$ and $\mathbf{M}_\mathbf{B}$ models, 1 instance or 3 instances result in state \textit{undbounded or infeasible}. These instances all belong to the set of 22 unsatisfiable instances. We discuss this result in more depth at the end of this section when we summarize our findings.





Table~\ref{result-omt} summarizes the results for the  OMT solvers Z3 and OptiMathSAT using their specific variants of the two different SMT2 models derived from the models $\mathbf{M}_\mathbf{Z_1}$ and $\mathbf{M}_\mathbf{Z_2}$. We show  the results for the OMT strategy for OptiMathSAT and the MaxRes strategy for Z3.
 
\begin{figure}[htb]
\begin{center}	
{\footnotesize
\begin{tabular}{|l|r|r|r|r|}  \hline 
& \multicolumn{2}{|c|}{Z3 MaxRes} & \multicolumn{2}{|c|}{OptiMathSAT OMT} \\ 
Solver State & SMT2$_{1-Z3}$ & SMT2$_{2-Z3}$ &
SMT2$_{1-OM}$  & SMT2$_{2-OM}$ \\  \hline	
optimal  				& 94 	& 96	&  	26 & 26 \\     	
suboptimal      		& 0		&  0	& 0 & 0 \\ 		 
unsolved  				& 160	& 158	& 	230 & 230	\\ 
unsatisfiable   		& 22	& 22	& 	 22 & 22	\\ 	
undefined    			& 2	 	& 2 	& 2& 2 \\  \hline			
\textbf{TOTAL Solved}	& 94	& 	96    	& 26 & 26\\ \hline		
\end{tabular}}
\caption{\label{result-omt}Performance of OMT solvers on entire benchmark set.}
\end{center}	
\end{figure}
	
As we defined in Table~\ref{solverstates}, optimally solved instances are determined for Z3 by comparing its solution costs to costs of known optimal solutions found by other solvers. Interestingly, Z3 only finds cost-optimal solutions or leaves an instance unsolved. From the solver state itself, all solutions found are marked as suboptimal by Z3. OptiMathSAT finds optimal solutions for 26 instances, but cannot solve any other instances. Both solvers also identify the 22 unsatisfiable instances easily. For the instances R001 and R002, the SMT2 tool chain has problems in generating correct files causing an error returned by both solvers, which we count for the undefined state. However, the problem here lies less with the solver, but rather with the instance file generation, which seems to be incomplete for these instances. Both solvers have difficulties scaling to larger instances. The permutation length of the largest instance that Z3 can solve is 36 and the average permutation length is only 17. The average number of atomic constraints for these instances is 38 and 9 for the disjunctive constraints. Only 1 instance solved by Z3 contains more than 222 atomic constraints, only 5 instances contain between 100 and 160 atomic constraints. For OptiMathSAT, the parameters are significantly lower. The average permutation length of the 26 solved instances is 7 and the largest solved instance has permutation length 12. This largest instance also contains the maximum number of 35 atomic and 13 disjunctive constraints. On the average, the instances solved by OptiMathSAT contain only 8 atomic and 2 disjunctive constraints.


\subsection{Impact of Tuning Search Strategies on Solver Performance}
\label{section:tuning}

We ran a few experiments to investigate if the performance of constraint solvers can be further improved by tuning search strategies. For Cplex CP, we investigated the influence of extended \textit{inference level setting}s, which allow the constraint solver to control the strength of domain reduction that it can achieve on the constraint variables by performing more or less constraint propagation. We set three inference levels to the value \textit{extended}: \textit{default inference level},  \textit{precedence inference level}, and \textit{allDifferent inference level}.

For the Chuffed solver, we experimented with different search annotations in MiniZinc.  We first experimented with an annotation on the \textit{alldifferent} constraint in the $\mathbf{M}_\mathbf{Z_1}$ and $\mathbf{M}_\mathbf{Z_2}$ models to use \textit{bound}s or \textit{domain} propagation :

{\scriptsize
	\begin{verbatim}
constraint all_different(pfc)::bounds;
constraint all_different(pfc)::domain;
\end{verbatim}
}
However, this annotation had no  impact on the number of instances that Chuffed can solve.  Chuffed can find a suboptimal solution for one more instance only on the $\mathbf{M}_\mathbf{Z_2}$ model using domain propagation. Solution costs even increased slightly. We therefore abandoned this annotation. 

We then experimented with various combinations of search annotations in order to control how the Chuffed solver conducts its variable choices and how it selects the  domain values for a variable, which seemed appropriate for the CTW domain. None of the combinations had a relevant impact on the number of instances solved by Chuffed, but sometimes solutions of lower cost are found. Table~\ref{search-anno} summarizes the relative changes in costs on the same subset of 208 (43 artificial and 165 real-world) instances where Chuffed finds an optimal or suboptimal solution in its default configuration (costs are set to 100 \%) or when using any of the different combinations of search annotations. For the choice how to constrain a variable, \textit{indomain\_split\_random}, which assigns a random value from the variable’s domain, works best, whereas the variable choice settings lead to no clear picture.  For the $\mathbf{M}_\mathbf{Z_1}$ model, the \textit{first\_fail} strategy (choose the variable with the smallest domain size) works best and for the 
$\mathbf{M}_\mathbf{Z_2}$ model, the \textit{most\_constrained}  strategy (choose the variable with the smallest domain, breaking ties using the number of constraints) works best.

\begin{figure}[htb]
\begin{center}	
{\footnotesize
\begin{tabular}{|l|l|r|r|r|r|r|}  \hline 
& & \multicolumn{5}{|c|}{\textbf{Variable Choice}}  \\ 
    & \textbf{Variable Value Choice} &  dom\_w\_deg &	 first\_fail &	 impact &	 most\_constrained 	& occurence \\ \hline	
 $\mathbf{M}_\mathbf{Z_1}$ & indomain\_min           & 99.91 & 99.87 & 99.66 & 99.99 & 100.00 \\\hline	
 $\mathbf{M}_\mathbf{Z_1}$ & indomain\_split\_random & 97.71 & \textbf{97.69} & 97.73 & 98.28 & 97.84 \\\hline	    	
 $\mathbf{M}_\mathbf{Z_2}$ &  indomain\_min   	      & 99.90 & 99.83 & 100.00 & 99.94 & 99.80 \\ \hline		 
 $\mathbf{M}_\mathbf{Z_2}$ & indomain\_split\_random & 97.66 &	97.65 &	97.85 &	\textbf{97.50} &	97.55 \\ \hline		
\end{tabular}}
\caption{\label{search-anno}Impact of different combinations of search annotations  on solution costs in Chuffed compared to Chuffed's performance without search annotations taken as 100\% baseline. Best values are marked in bold.}
\end{center}	
\end{figure}

Google OR-Tools aborts search with a message that the search annotation is unsupported when we invoked it on the regenerated flatzinc files from the MiniZinc models containing search annotations. We therefore tested it on a dual model that we derived from the  $\mathbf{M}_\mathbf{Z_1-NoAbs}$ model by adding the \textit{cfp}  array and the corresponding \textit{alldifferent} and \textit{channeling} constraints:

{\scriptsize
	\begin{verbatim}
array[Positions] of var Cavities: cfp;
constraint all_different(cfp);
constraint inverse(pfc,cfp);
	\end{verbatim}
}

We also fed this model into Chuffed and added the best working search annotation to this model:

{\scriptsize
	\begin{verbatim}
	::int_search(pfc, first_fail,indomain_split_random)
	\end{verbatim}
}

Table~\ref{cp-tuning} summarizes the results for Cplex CP with extended inference level settings on the dual model $\mathbf{M}_\mathbf{C}$ and Google OR-Tools and Chuffed on the dual extension of the $\mathbf{M}_\mathbf{Z_1-NoAbs}$ model. Chuffed is run with search annotations and without \textit{freesearch}, \ie without the additional option to deviate from the annotated search strategy. The results seem to confirm the observation made in~\cite{smith} that using a dual model works best when modeling permutation problems.

With extended inference level settings, Cplex CP finds optimal solutions for 146 instances. This set contains the same 138 instances that Cplex CP could solve optimally using the default inference level settings, but not the instance R009, for which only a suboptimal solution is found. In addition, Cplex CP finds another 56 solutions of minimal costs, but cannot prove these solutions as optimal. Without extended inference level settings, Cplex CP found 54 cost-minimal solutions. It can now prove some of these solutions to be optimal. The number of cost-minimal solutions thus grows from 193 to 202. Total solution costs for the 256 solved instances is reduced by 3\%.

All three solvers can solve the largest number of instances using the dual modeling approach.  Table~\ref{cp-tuning} also compares the cost and runtimes on the subset of 238 instances, for which all three solvers can find solutions. Note that this subset contains different instances than the set of 218 instances taken as basis for the cost and runtime comparisons in Figure~\ref{costs-cpsat}. Out of the previously solved 218 instances, all solvers now solve 208 of them and there are 10 instances, which are not solved by Chuffed or OR-Tools using the dual model although they found solutions for these instances previously using the other models. OR-Tools again shows the fastest performance, which is explained by the high number of optimal solutions that it finds and for which it rarely exhausts the 5 minute time limit. Among all three solves, Cplex CP finds solutions of significantly lower costs for this subset of 238 instances.

\begin{figure}[htb]
\begin{center}	
{\footnotesize
\begin{tabular}{|l|r|r|r|r|}  \hline 
& \multicolumn{1}{|c|}{Cplex CP} & \multicolumn{1}{|c|}{Chuffed} & \multicolumn{1}{|c|}{OR-Tools } \\ 
Solver State &  \multicolumn{1}{|l|}{Extended Inference}  &  \multicolumn{1}{|l|}{Dual Model +}       & \multicolumn{1}{|l|}{Dual Model}\\ 	
             &  \multicolumn{1}{|l|}{Level Settings}      &  \multicolumn{1}{|l|}{Search Annotation +}  &  \multicolumn{1}{|l|}{+}\\
       &  \multicolumn{1}{|l|}{ }      &  \multicolumn{1}{|l|}{No freesearch}  & No freesearch \\\hline       
             
optimal  				& 146 (+56) 	& 116	& 225			\\     	
suboptimal  			& 110 			& 125 	& 20	\\ 		 
unsolved  				& 0 			& 14 	& 9	\\ 
unsatisfiable   		& 22 			& 22	& 22	\\  	
undefined    			& 0 			& 1 	& 2			\\  \hline				
\textbf{TOTAL Solved}	& 256			& 241 	& 245		\\ \hline	
total costs (238 instances) & 899,333,199 &  1,855,447,444 	&   1,185,071,400  \\ 
relative costs (in \%)	& 	76 & 157	& 100  \\ \hline
total runtimes (in s)	& 30,624	& 40,365 &	14,316 \\ 
relative runtimes (in \%) & 214 & 282	 & 	100 \\ \hline

\end{tabular}
}
\caption{\label{cp-tuning}Performance of constraint solvers with tuned search strategies using the dual CTW model. Relative cost and runtime comparisons are calculated taking the values of OR-Tools as 100 \%.}
	\end{center}	
\end{figure}


Finally, we tested Z3 on the benchmark set with its other solvers WMax~\cite{wmax1,wmax2} and PD-MaxRes~\cite{pd-maxres} and compared them to the MaxRes strategy. All strategies compare slightly better on the $\mathbf{SMT2_{1-Z3}}$ model, but there are only small differences between the strategies, see Table~\ref{result-z3}. As with the MaxRes strategy, Z3 using Wmax or PD-MaxRes either returns an optimal solution or leaves a problem instance unsolved. We also tried to run OptiMathSAT with its MaxRes strategy, but got an error message on all generated SMT2 files indicating that it had problems extracting the objective function when using this strategy.

\begin{figure}[htb]
\begin{center}	
{\footnotesize
\begin{tabular}{|l|r|r|r|r|r|r|}  \hline 
& \multicolumn{2}{|c|}{Z3 MaxRes} & \multicolumn{2}{|c|}{Z3 PD-MaxRes} & \multicolumn{2}{|c|}{Z3 WMax}\\ 
Solver State & $\mathbf{SMT2_{1-Z3}}$ & $\mathbf{SMT2_{2-Z3}}$ &
$\mathbf{SMT2_{1-Z3}}$  & $\mathbf{SMT2_{2-Z3}}$ &
$\mathbf{SMT2_{1-Z3}}$  & $\mathbf{SMT2_{2-Z3}}$ \\  \hline	
optimal  				& 94 	& 96	& 94 	& 97    & 91  	& 97\\     	
suboptimal      		& 0	 	&  0	& 0  	&  0    &  0    & 0\\ 		 
unsolved  				& 160	& 158	& 160	& 157	& 163 	& 157\\ 
unsatisfiable   		& 22	& 22	& 22	& 22	& 22 	&  22 \\ 	
undefined    			& 2		& 2 	& 2	    &  2	& 2     & 2 \\  \hline			
\textbf{TOTAL Solved}	& 94	& 96	&  94   & 97	&  	91  & 97\\ \hline		
\end{tabular}}
\caption{\label{result-z3}Comparison of different Z3 strategies on entire benchmark set.}
\end{center}	
\end{figure}

\subsection{Summary of Findings and Research Challenges}
\label{findings}

Our empirical analysis illustrated the varying performance of the various solvers on the CTW benchmark set. In particular, modern constraint solvers showed impressive results and notably IBM Cplex CP and Google OR-Tools CP-SAT solvers excel in the tests. For some solvers,  their performance  can be further tuned by setting options in the search strategies, however, we believe that the future will be in automatic, rather than human-provided search strategy selection. In addition, we are convinced that tuning/rewriting the models has further potential, in particular, for improving the performance of MIP solvers. In the following, we summarize our findings, derive research challenges for constraint solvers, and discuss some issues for further maturing solvers towards complex real-world applications.

\paragraph{Simplifying Benchmarking Experiments} We invested about one person year into the empirical testing of the solvers, which turned out be much more complex than expected. In particular, the manual rewriting of models for the MIP solvers and the semi-automatic generation of SMT2 files were time-consuming and error-prone steps requiring to write substantial pieces of software. Having a software environment in place, which allowed us to integrate all solvers and in particular also to automatically write and analyze log files as well as to validate all solver solutions was instrumental to obtain reproducible results. Our work also emphasizes the need for a unified modeling language that would ease the exchange of models and data between different algorithms. Furthermore, having a standardized output interface in place to extract results and optimization costs would lower the benchmarking burden. Furthermore, we believe that following a modeling approach, which keeps models and instance data separately, provides an easier-to-access interface to solvers. 

\paragraph{Undefined Solver States} 
At the beginning of the experimental series, we defined a mapping of individual solver states to a common set of states, recall Table~\ref{solverstates}. States, which a solver returned and which were not part of our mapping, are mapped to a value of "undefined". Interestingly, we obtained more such states than expected. Several solvers have issues with instances R001 and R002. Instance R001 contains no cables and the empty permutation is the solution, instance R002 contains one one-sided cable, \ie a single job and no constraints, but it is solvable with the permutation containing this single job. Independently of the model used with the solver, solvers returned states as summarized in Table~\ref{result-r}.

\begin{figure}[htb]
\begin{center}	
{\footnotesize
\begin{tabular}{|l|l|l|r|r|r|r|}  \hline 	
Instance & k & b & Chuffed 	& OR-Tools 	& Z3/OptiMathSAT  	& Gurobi   \\  \hline	
R001  	 & 0 & 0 & Error  			& empty log 	& empty log   	& 	infeasible model	 \\     	
R002     & 1 & 0 & valid solution  	& empty log  	& empty log  	&  valid solution    	 \\ \hline		
		\end{tabular}}
		\caption{\label{result-r}Results of solvers on pathological instances.}
	\end{center}	
\end{figure}

The Chuffed solver returns an error and no solution for instance R001 on both of its models.  The OR-Tools CP-SAT solver works on instances R001 and R002 for about 1.5 seconds and then returns empty logs without a solution.  Z3 and OptiMathSAT also fail on these 2 instances. One possible explanation could be problems in the generation of the FlatZinc and SMT2 file generation, notably for instance R001. For example, the .FZN file for R001 contains an array with bounds set to 1..0. The generation of the SMT2 files for instances R001 and R002 generates an error message "error: failed to generate SMT-LIB formula" thrown by the OptiMathSAT binary. The Gurobi MIP solver reports an infeasible model for instance R001. 
 
The undefined state for the Cplex MIP solver on various instances is caused by a state of "unbounded or infeasible" returned by the solver. The result was surprising as none of these instances is infeasible and our models are not unbounded. Upon closer inspection of the behavior of the solver, we located the reasons for this state by the presolve strategy applied by Cplex MIP. Switching off presolving allows the solver to either find a suboptimal solution or run into the time-limit without finding a solution. This behavior is known for problems, which are "borderline infeasible".\footnote{See also \url{https://www.ibm.com/support/pages/turning-cplex-presolve-or-gives-inconsistent-results} and \url{https://www.ibm.com/support/knowledgecenter/SSSA5P_12.10.0/ilog.odms.cplex.help/CPLEX/Parameters/topics/PreInd.html}
.} Further investigating the borderline infeasibility of some our instances, which is likely caused by the high number of constraints, would definitely be an interesting avenue for future research.

\paragraph{Deviations in Solution Costs} 

Our validation software ensures that all constraints are satisfied by a solution returned by a solver and it also recalculates the costs of the criteria $S$, $M$, $L$, and $N$ as well as the value of the overall objective. All cost values shown in the figures are based on these recalculations. In case of the OMT solvers, we added the optimization criteria $S,L,M,N$ as output variables in addition to the overall value of the optimization objective, but still had problems in accessing the value of single criteria such as $S$ and $L$ for example, however, we did not investigate this issue further as our recalculations were available. 

Figure~\ref{costs-cpsat} showed some cost deviations for the Cplex CP constraint solver, where the optimal solutions for 4 instances violate more soft atomic constraints (criterion $N$) than optimal solutions found by other solvers on same instances: A001 (+2), R126 (+6), R127 (+3), and R128 (+6). Cplex indeed returns these higher-cost solutions as having the best objective. For each instance, it shows a best bound having a lower value and these solutions are within the default optimality tolerance, which is $1.0 \; e^{-9}$. For example, for instance R128 an optimal solution of costs 129,729 is computed and the effective tolerance is 12.9729. 

Furthermore, a deviation in the computation of the $M$ criterion by Chuffed and on the $N$ and $M$ criterion by the Cplex MIP solver was detected by our validation. On the $M$ criterion, which captures how long a cable is kept in storage, our validated values are sometimes higher by 1 than those reported by the solver, but again only on very few instances. Apparently, the interpretation of how the value of $M$ and $N$ is computed differs slightly in our validation code from how the solvers interpret the specification of the corresponding decision expression, but only for very few instances. These minor deviations show that first, revalidating solution costs is important when solutions from different algorithms are compared with each other. Second, the specification of optimization objectives is a challenging task and it also heavily depends on how a problem is modeled. Having good support available in modeling languages and solvers to formulate and test optimization objectives is highly desirable in particular from an application-oriented perspective.


\paragraph{Algorithmic Insights For Improving Models} 

We presented several modeling variants for the CTW problem, which were all created manually by applying different modeling approaches. The development process of a model proceeds over many iterations and is often an error-prone process. Quite often it can happen that incorrect formulations of constraints render an instance unsatisfiable. Although constraint solvers can quickly identify minimal conflict sets of constraints, finding and removing the root cause of an inconsistency is not straightforward. Tool support to further analyze inconsistencies would be more than desirable and be another promising avenue for further research. Furthermore, feedback from solvers would be desirable that helps in understanding what parts of a model make it difficult to solve. From a user's perspective, a model should be as compact and easy to understand as possible. From a solver's perspective, the model should allow for maximum constraint propagation for example. Similarly to determining the best possible search strategy automatically, it would be highly desirable if solvers could automatically compile/rewrite models into more effective representations. Some early work exists~\cite{heinz,chu,nightingale,rendl} and we argue that much more can be done here.

\paragraph{Recognizing Hard and Easy Instances} 

The CTW benchmark set comprises instances of varying difficulty. The sum of constraints measure introduced in Section~\ref{benchmark} appears to be a good first indicator of the difficulty of each instance. In Figure \ref{figure:quartilesSum} we give an overview of the first and third quartile of the constraint sum for each solver and solver state. The entries in the table correspond to the results of the best model for each solver. While a correlation can be observed between the constraint sum and the solving state of each solver, the constraint sum does not give any indication on which types of constraints contribute the most to the difficulty of the instance. From a theoretical point of view, better understanding the phase transitions~\cite{cheeseman} of this benchmark set is an interesting research problem. From a practical point of view, better understanding the hardness and possible solution quality is desirable. Cplex CP is the only solver which finds solutions for all instances in the benchmark set, but it reports for example a gap of over 98 \% for the large instances A70 to A73 with permutation length between 190 and 198 and between 7,000 and 10,000 atomic constraints. In the CTW application, the number of generated constraints can be influenced by choosing a different layout of harnesses on the palette. If one could know better which subsets of constraints render an instance difficult, insights into how to modify a layout seem to be within reach. 

\begin{figure}[htb]
\begin{center}	
{\footnotesize
\begin{tabular}{|l|r|r|r|r|r|r|r|}\hline 
              	& Chuffed		&  Cplex CP		& OR Tools &	 Cplex MIP  & Gurobi	& Z3  PDMaxRes	& OptiMathSAT    \\ 
				& $\mathbf{M}_\mathbf{Z_1}$     &  $\mathbf{M}_\mathbf{C}$ + Tuning  &  Dual Model & $\mathbf{M}_\mathbf{I}$
				&  $\mathbf{M}_\mathbf{B C\sharp}$ &      $\mathbf{SMT2_{2}}$      &  $\mathbf{SMT2_{1}}$        \\  \hline				
optimal Q1		&49.50			&48.5			&	71.00		&	42.00		& 44.50		& 30.00 	&7.50			\\ \hline
optimal Q3		&203.25 		&207.00			&	572.00		&	158.00		& 157.75	& 96.00		&23.50			\\ \hline
suboptimal Q1	&410.00	 		&415.25			&	1072.75		&	274.00	    & 292.00	& 			&				\\ \hline
suboptimal Q3	&827.00			&1533.00		&   3481.00  	&	682.00		& 638.50    & 			&				\\ \hline
unsolved Q1	&1551.75	 	&				&	9445.00 	&	755.25		& 576.25    & 320.00	&118.75			\\ \hline
unsolved Q3	&9267.25	 	&				&	11171.00	&  7016.50		& 1551.5	& 902.00	&751.75			\\ \hline
\end{tabular}
\caption{\label{figure:quartilesSum} First (Q1) and third (Q3) quartile of the constraint sum for each solver. No entry in a row means that no data was available for this solver state as the solver found for example only optimal, but no suboptimal solutions such as in the case of the OMT solvers. }}	
\end{center}	
\end{figure}

\section{Conclusion}
\label{conc}
We discuss the problem of cable tree  wiring (CTW), which we position as a variant of a traveling sales person problem with atomic, soft atomic, and disjuntive precedence constraints, direct successor constraints as well as tour-dependent edge costs. The CTW problem can also be considered as the first known representative of the coupled task scheduling problem with soft constraints and as a new variant of the pickup and delivery TSP. Using the relationships to these known problems, we prove the NP-hardness of various subclasses of the CTW problem and also show that certain restrictions of the various constraint sets can make the problem solvable in polynomial time. In addition, we identify interesting subclasses of the problem for which the complexity is open. We also discuss the constraint sum parameter as a promising predictor for the difficulty of solving an instance. 

We present a benchmark set of 278 real-world and artificial instances and compare state-of-the-art constraint, mixed-integer, optimization modulo theory solvers on this set using also different variants of how the problem can be modeled. Given our modeling variants, in particular IBM Cplex CP and the Google OR-Tools CP-SAT solver showed impressive results, with Cplex being the only tested solver to find solutions for each instance in the benchmark set and OR-Tools finding more optimal solutions than any of the other solvers we tested. Note however, that models had to be rewritten for different solvers and we believe that tuning/rewriting the models has further potential, in particular, for improving the performance of MIP solvers.
Our results demonstrate the remarkable progress made over recent years, in particular in the field of constraint and CP-SAT solvers, and also raises several interesting questions for future research.

\bigskip

\noindent
\textbf{Acknowledgment} This work was partially supported by the Swiss Innovation Agency innosuisse. We thank Stefan Bucheli, Beat Estermann, Roland Liem, Georg Moravitz, Kurt Ulrich,  and Zeta technicians from  Komax AG for their support, fruitful cooperation, and access to this interesting data set. 
Deep thanks goes to Bernhard Nebel, Albert-Ludwigs-University Freiburg, for a fruitful discussion on the complexity of the CTW problem and to Guido Tack, Monash University, for feedback on an earlier version of this paper and for contributing one of the models.

\bibliographystyle{plain}
\bibliography{cabletree}

\newpage

\appendix

\section{Appendix: Details of Models $\mathbf{M}_\mathbf{Z_1}$, $\mathbf{M}_\mathbf{Z_2}$, $\mathbf{M}_\mathbf{I}$, $\mathbf{M}_\mathbf{D}$, $\mathbf{M}_\mathbf{B}$}

Starting from the  "native" model $\mathbf{M}_\mathbf{C}$, we manually rewrote this model into the models $\mathbf{M}_\mathbf{Z_1}$ in the language MiniZinc and into the model $\mathbf{M}_\mathbf{I}$ in the OPL language variant supported by the Cplex MIP solver. Both models do not use the dual representation of the permutation sequence. The MiniZinc model then provided the starting point for further transformations as described in Figure~\ref{fig:model-overview3b} to feed our instances into various other solvers. The model $\mathbf{M}_\mathbf{D}$ carries the dual modeling approach over to the MIP solvers, however, it turned out that these solvers are rather negatively impacted by the dual model, whereas this model was instrumental for the Cplex constraint solver to scale to larger CTW instances. To complement our own approach of modeling the CTW problem, we were also interested in testing solvers on different modeling approaches. This led to the model $\mathbf{M}_\mathbf{B}$ for the MIP solvers and the MiniZinc model $\mathbf{M}_\mathbf{Z_2}$ for the constraint and OMT solvers. In the following, we give an overview over these models. 

\subsection{The constraint models $\mathbf{M}_\mathbf{Z_1}$ and $\mathbf{M}_\mathbf{Z_2}$ in MiniZinc Language}

In contrast to the Cplex OPL language, the MiniZinc language does not provide possibilities to represent tuples. We begin by declaring a number of integer parameters for our models.  OPL ranges are translated into sets of integers in MiniZinc.

{\scriptsize
	\begin{verbatim}
	int: k;
	set of int: Positions = 1..k;
	set of int: Cavities = Positions;
	int: b;
	set of int: CavityPairs = 1..2*b;
	set of int: CableStarts = 1..b;
	\end{verbatim}
}

Constraint types are represented using arrays. The arguments of the arrays are the cavities between which the constraint must hold. For disjunctive constraints, a 2-dimensional array is used. 

{\scriptsize
	\begin{verbatim}
	array[int,int] of Cavities: AtomicConstraints;
	array[int,int] of Cavities: DisjunctiveConstraints;
	array[int] of Cavities: DirectSuccessors;
	array[int,int] of Cavities: SoftAtomicConstraints;
	\end{verbatim}
}

We do not use the dual representation of {\em pfc} and {\em cfp}, but only introduce the {\em position\_for\_cavity (pfc)} to represent the desired permutation sequence. MiniZinc supports various global constraints, which are included into the model to express the {\em all\_different} constraint over the elements of this array in a straightforward way.

{\scriptsize
	\begin{verbatim}
	array[Cavities] of var Positions: pfc;
	include "globals.mzn";
	constraint all_different(pfc);
	\end{verbatim}
}

Constraints are declared using the keyword {\em constraint} and make use of the {\em index\_set} notation in MiniZinc. The formulation of the disjunctive constraint also makes use of the special case when the cavity on the left-hand side of the precedence relation is the same in both disjuncts similar to the Cplex $\mathbf{M}_\mathbf{C}$ model. 

{\scriptsize
	\begin{verbatim}
	constraint 
	forall (i in index_set_1of2(AtomicConstraints)) 
	(pfc[AtomicConstraints[i,1]] < pfc[AtomicConstraints[i,2]]);
	
	constraint 
	forall (i in index_set_1of2(DisjunctiveConstraints)) 
	(( pfc[DisjunctiveConstraints[i,1]] < pfc[DisjunctiveConstraints[i,2]] 
	\/
	pfc[DisjunctiveConstraints[i,3]] < pfc[DisjunctiveConstraints[i,4]]
	) 
	/\
	if DisjunctiveConstraints[i,1]==DisjunctiveConstraints[i,3] then
	max(pfc[DisjunctiveConstraints[i,2]], pfc[DisjunctiveConstraints[i,4]]) 
	> pfc[DisjunctiveConstraints[i,1]]
	else true endif);
	
	constraint 
	forall (i in index_set(DirectSuccessors)) 
	(if DirectSuccessors[i]<= b 
	then pfc[DirectSuccessors[i]] < pfc[DirectSuccessors[i]+b] 
	-> pfc[DirectSuccessors[i]] +1 = pfc[DirectSuccessors[i]+b]
	else pfc[DirectSuccessors[i]] < pfc[DirectSuccessors[i]-b] 
	-> pfc[DirectSuccessors[i]] +1 = pfc[DirectSuccessors[i]-b]
	endif);
	\end{verbatim}
}

The various decision variables for the optimization criteria as well as the objective function {\em obj} are defined as follows: 

{\scriptsize
	\begin{verbatim}
	var int: S = 
	if b=0  then 0 
	else sum(i in CableStarts) (abs(pfc[i]-pfc[i+b]) > 1) 
	endif;
	
	var int: M = 
	if b=0 then 0 
	else (max(i in CavityPairs) 
	(sum(j in CavityPairs where j<=b) 
	(pfc[j] < pfc[i] /\ pfc[i] < pfc[j+b])
	+sum(j in CavityPairs where j>b) 
	(pfc[j] < pfc[i] /\ pfc[i] < pfc[j-b])))
	endif;
	
	var int: L = 
	if b=0 then 0 
	else max(i in CableStarts) (abs(pfc[i]-pfc[i+b])-1)
	endif;
	
	var int: N = 
	sum(i in index_set_1of2(SoftAtomicConstraints)) 
	(pfc[SoftAtomicConstraints[i,1]] > pfc[SoftAtomicConstraints[i,2]]);
	
	var int: obj = S*pow(k,3)+M*pow(k,2)+L*pow(k,1)+N;
	\end{verbatim}
}

A different modeling approach is adopted for the $\mathbf{M}_\mathbf{Z_2}$ model, which was contributed by Guido Tack, Monash University. The disjunctive constraint can be rewritten using an array of booleans and an additional constraint over this array can be added. The array captures the truth value of the two disjuncts in the disjunctive constraint and the constraint states that at least one of the disjuncts must be true for the disjunctive constraint to be satisfied. 

{\scriptsize
	\begin{verbatim}
	array[int] of var bool: disj =
	[ pfc[DisjunctiveConstraints[i,2*j+1]] < pfc[DisjunctiveConstraints[i,2*j+2]] 
	| i in index_set_1of2(DisjunctiveConstraints), j in 0..1
	];
	
	constraint forall (i in 1..length(disj) div 2) (disj[(i-1)*2+1] \/ disj[(i-1)*2+2]);
	\end{verbatim}
}

An array of booleans is used to capture the values of the decision variables $S$ and $N$ by introducing two additional variables {\em vio\_abs} and {\em vio} and then representing the variables as sums over these boolean array values. The {\em vio\_abs} array captures whether a cable end must be stored or not, based on whether the "other end" of the cable is plugged before or after the "cablestart". The {\em vio} array captures if the positions of two cavities in the permutation sequence satisfy or violate a given soft atomic constraint over these two cavitities.

{\scriptsize
	\begin{verbatim}
	array[int] of var bool: vio_abs = 
	[ pfc[i]-pfc[i+b] > 1 \/ pfc[i+b]-pfc[i] > 1 | i in CableStarts];
	
	var int: S = sum(vio_abs);
	
	array[int] of var bool: vio = 
	[pfc[SoftAtomicConstraints[i,1]] >= pfc[SoftAtomicConstraints[i,2]] 
	| i in index_set_1of2(SoftAtomicConstraints)];
	
	var int: N = sum(vio);
	\end{verbatim}
}

\subsection{The Mixed-integer Programming Models  $\mathbf{M}_\mathbf{I}$ and $\mathbf{M}_\mathbf{D}$}

The mixed-integer programming model $M_I$ was manually rewritten from the $M_C$ model in OPL without any sophisticated reformulation of the constraints as for example discussed in~\cite{refalo} in order to keep the model close to the constraint model $M_C$.
The model is based on the {\em pfc} permutation, does not make use of the dual {\em cfp} representation, and thus needs no channeling constraint. The {\em allDifferent} constraint is replaced by pairwise inequalities over cavity numbers assigned to {\em pfc} positions. All constraints are represented using tubples. Precedence constraints are reformulated as inequalities and direct successors are formulated in equation form. 

{\scriptsize	
	\begin{verbatim}	
	int k  = ...; // number of cavities, permutation length
	int b = ...;  // number of wired cavity pairs
	
	range Cavities = 1..k;
	range Cablestarts = 1 ..b;
	
	tuple Atomic{ int cbefore; int cafter;};
	
	// disjunctive constraints
	tuple Disjun{ int c1before; int c1after; int c2before; int c2after;  };
	
	{Atomic} AtomicConstraints = ...;
	{Atomic} SoftAtomicConstraints = ...;
	{Disjun} DisjunctiveConstraints = ...;	
	{int} DirectSuccessors = ...;
		
	range Positions = 1..k;
	range CavityPairs = 1..2*b ;
	
	dvar int pfc[Cavities] in Positions;
	
	
	//alldifferent
	forall(i, j in Cavities: i != j) pfc[i] != pfc[j];
	
	//atomic constraints
	forall(c in AtomicConstraints)
	pfc[c.cbefore] - pfc[c.cafter]  <= 0;        
	
	//disjunctive constraints
	forall(d in DisjunctiveConstraints)
	(pfc[d.c1before] -  pfc[d.c1after] <= 0 ||  pfc[d.c2before] -  pfc[d.c2after]  <= 0 );
	
	// direct successor constraints
	forall (i in DirectSuccessors: i <= b)  
	pfc[i] - pfc[i+b] <= 0 =>  pfc[i+b] == pfc[i] + 1;
	
	forall (i in DirectSuccessors: i > b)  
	pfc[i] - pfc[i-b] <= 0 =>  pfc[i-b] == pfc[i] + 1;
	\end{verbatim}
} 

The optimization criteria are rewritten similarly and the objective function remains unchanged. Note that the absolute-function cannot be used for the criterion $L$ as it returns a float value, which would make the problem unbounded. 

{\scriptsize	
	\begin{verbatim}	
	dexpr int S = ((b == 0) ? 0: sum(j in Cavities: 1<= j<= b) 
	(maxl(pfc[j], pfc[j+b])-minl(pfc[j], pfc[j+b]) >= 2));
	
    dexpr int L = ((b == 0) ? 0 : max (j in Cablestarts) 
     (maxl(pfc[j] - pfc[j+b], pfc[j+b] - pfc[j]) - 1));
	
	dexpr int M = ((b == 0) ? 0   : (max (i in CavityPairs)
	(sum(j in CavityPairs: j<=b) (pfc[j] <= (pfc[i] - 1) && 
	pfc[i] <= (pfc[j+b] - 1)) 
	+ sum(j in CavityPairs: j>b) (pfc[j] <= (pfc[i] - 1) && 
	pfc[i] <= (pfc[j-b] - 1)))));
	
	dexpr int N =	sum(s in SoftAtomicConstraints)
	(pfc[s.cafter] - pfc[s.cbefore] <= 0);
	
	minimize S * pow(k, 3) + M * pow(k, 2) + L * pow(k, 1) + N;	
	\end{verbatim}
}

The model $M_D$ extends the model $M_I$ with the dual {\em cfp} representation, the inequality constraint for {\em cfp} and two channeling constraints:

{\scriptsize
	\begin{verbatim}
	// position for chamber
	dvar int pfc[Cavities] in Positions;
	//chamber for position
	dvar int cfp[Positions] in Cavities;
	
	//alldifferent
	forall(i, j in Cavities: i != j) pfc[i] != pfc[j];
    forall(i, j in Positions: i != j) cfp[i] != cfp[j];

	
	//duality (channeling constraint)
	forall(j in Cavities, p in Positions)
	pfc[j] == p =>  cfp[p] == j;
	
	forall(j in Cavities, p in Positions)
	cfp[p] == j =>  pfc[j] == p;	
	\end{verbatim}
}

\subsection{The Mixed-integer Programming Model  $\mathbf{M}_\mathbf{B}$}

The model $M_B$ makes use of big-M reformulations for disjunctive constraints~\cite{ruiz} using additional decision variables and non-integer (floating point) optimization costs. The value of the big-M constant has to be chosen sufficiently big to prevent the problem from becoming unsolvable. As a rule of thumb, the big-M constant should be at least 100 times larger than the largest value of any of the variables. However, it should not be too large, because otherwise numerical instability and round-off errors can occur. Using an arbitrarily large M value also expands the feasible region of the LP relaxation unnecessarily and results in longer runtimes, see~\cite{camm}. We decided to set \verb+int bigM=k*100+ as smaller or larger values led to incorrect values for some optimization criteria.

The following model $\mathbf{M}_\mathbf{B}$ for the Cplex MIP solver uses an upper triangular matrix of booleans that indicates if a cavity is wired before another:

{\scriptsize
	\begin{verbatim}
	dvar boolean lt[Cavities][Cavities];
	\end{verbatim}
}

If $lt[i,j] = 1$, then {\em pfc[i] $<$ pfc[j]}, otherwise {\em pfc[i] $>$ pfc[j]}. Only the upper triangular matrix is defined, \ie  values are only defined for $i<j$. To find out if {\em pfc[i] $<$ pfc[j]} when $i>j$, the value of $lt[j,i]$ is determined and inverted. The diagonal of the $lt$ matrix is undefined, because a cavity cannot be placed before or after itself in the permutation.  A solution can be directly extracted from the $lt$ matrix as the subsequent example explains.

\begin{minipage}[c]{35mm}
	\begin{equation*}  
	lt = 
	\begin{bmatrix}
	& 1& 1& 1\\
	& & 0& 1\\
	&  & & 1\\
	&  &  & 
	\end{bmatrix}
	\end{equation*}
	
\end{minipage}
\begin{minipage}[c]{40mm}
	
	$pfc=1,3,2,4$  because 
	
	{\small 
		{\em pfc[1]$<$pfc[2]}, {\em pfc[1]$<$pfc[3]}, 
		{\em pfc[1]$<$pfc[4]}, {\em pfc[2]$>$pfc[3]}, 
		{\em pfc[2]$<$pfc[4]}, {\em pfc[3]$<$pfc[4]}}
\end{minipage}

\medskip

Atomic constraints are expressed as entries in the $lt$ matrix:

{\scriptsize
	\begin{verbatim}
	forall(a in AtomicConstraints)
	if (a.cbefore < a.cafter) lt[a.cbefore,a.cafter] == 1;
	else lt[a.cafter,a.cbefore] == 0;
	\end{verbatim}
}

The {\em allDifferent} constraint is expressed as inequalities over $lt$ matrix values stating that either {\em pfc[i]} or {\em pfc[j]} has to be larger than the other.

{\scriptsize
	\begin{verbatim}
	forall(ordered i,j in Cavities)
	pfc[i] - pfc[j] + 1 <= bigM * (1 - lt[i,j]);
	forall(ordered i,j in Cavities)
	pfc[j] - pfc[i] + 1 <= bigM * lt[i,j];
	\end{verbatim}
}

For example, if $lt[i,j]$ is equal to $0$ and therefore {\em pfc[i] $>$ pfc[j]}, the first constraint is always satisfied if the big-M constant is chosen big enough. The right term in the second constraint evaluates to $0$ and the left term has to be smaller or equal to $0$, which is only possible if the two values are different. Disjunctive constraints are formulated as inequality constraints. A satisfied disjunctive constraint is larger or equal to 1, because we sum up values for each atomic constraint in the disjunction from the $lt$ matrix. As only the upper triangular matrix of the $lt$ matrix is defined, case distinctions based on cavity indices have to be made and some values have to be inverted. With a fully populated matrix $lt$, each disjunctive constraint is expressed by the following conditional inequations. Note that for each
disjunctive constraint, only one of the if-clauses is satisfied.

{\scriptsize
	\begin{verbatim}
	forall(d in DisjunctiveConstraints){
	if (d.c1before < d.c1after && d.c2before < d.c2after)
	{lt[d.c1before,d.c1after] + lt[d.c2before,d.c2after] 
	>= 1;}
	if (d.c1before < d.c1after && d.c2before > d.c2after)
	{lt[d.c1before,d.c1after] + 1-lt[d.c2after,d.c2before] 
	>= 1;}
	if (d.c1before > d.c1after && d.c2before < d.c2after)
	{1-lt[d.c1after,d.c1before] + lt[d.c2before,d.c2after] 
	>= 1;}
	if (d.c1before > d.c1after && d.c2before > d.c2after)
	{1-lt[d.c1after,d.c1before] + 1-lt[d.c2after,d.c2before]
	>= 1;}}
	\end{verbatim}
}

Direct successor constraints also make a case distinction based on cavity indices. If $i$ is in {\em Directsuccessors}, then $i+b$ must follow immediately in the permutation, which means that the left-hand side of the condition must be equal to 0 if and only if {\em  pfc[i] $<$ pfc[i+b]}.

{\scriptsize
	\begin{verbatim}
	forall(i in Cablestarts: 
	i in DirectSuccessors || (i+b) in DirectSuccessors){
	if (i in DirectSuccessors)
	pfc[i+b] - pfc[i] - 1 <= bigM * (1 - lt[i,i+b]);
	if ((i+b) in DirectSuccessors)
	pfc[i] - pfc[i+b] - 1 <= bigM * lt[i,i+b];}	
	\end{verbatim}
}

Minimum and maximum positions of cavities in the permutation are stored in arrays {\em minimum} and {\em maximum}. Their values are set by constraints.

{\scriptsize
	\begin{verbatim}
	dvar float+ minimum[Cablestarts] in Positions;
	dvar float+ maximum[Cablestarts] in Positions;
	
	forall(i in Cablestarts) {
	minimum[i] - pfc[i]   <= 0;                        (1)
	minimum[i] - pfc[i+b] <= 0;                        (2) 
	pfc[i] - minimum[i]   <= bigM * (1-lt[i,i+b]);     (3) 
	pfc[i+b] - minimum[i] <= bigM * lt[i,i+b];         (4)
	maximum[i] == pfc[i]+pfc[i+b] - minimum[i];}       (5)
	\end{verbatim}
}

Constraints (1) and (2) ensure that the value of {\em minimum[i]} is larger or equal to {\em min(pfc[i],pfc[i+b])}. Constraints (3) and (4) ensure that the value of {\em minimum[i]} is smaller or equal to {\em min(pfc[i],pfc[i+b])}, depending on which of the cavities in a job pair occurs first in the permutation. If constraints (1) to (4) are satisfied, the value of {\em minimum[i]} is set correctly. Calculating the value of {\em maximum[i]} is then trivial using constraint (5).\\

If a job pair is interrupted, its entry in the array {\em cableIsStored} is set to 1, which happens if the difference between the {\em minimum} and {\em maximum} position of cavities from a job pair in the permutation is larger or equal than 2.

{\scriptsize
	\begin{verbatim}
	dvar boolean cableIsStored[Cablestarts];
	
	forall(p in Cablestarts) {
	2 - maximum[p] + minimum[p] <= bigM * (1-cableIsStored[p]);
	maximum[p] - minimum[p] - 1 <= bigM * cableIsStored[p];}
	\end{verbatim}
}

The variable {\em cableIsStoredAtPosition[i,t]} is equal to 1 if one cavity from the job pair $i$ occurs in the permutation before the position $t$, but the other not. Otherwise {\em cableIsStoredAtPosition[i,t]} is equal to 0. The variable {\em cableIsPluggedBefore[i,t]} is equal to 1 if both cavities of a job pair $i$ occur  in the permutation before position $t$. The variable {\em cableIsPluggedAfter[i,t]} is equal to 1 if both cavities of a job pair $i$ occur in the permutation after position $t$.

{\scriptsize
	\begin{verbatim}
	dvar boolean cableIsStoredAtPosition[Cablestarts,Positions];
	dvar boolean cableIsPluggedBefore[Cablestarts,Positions];
	dvar boolean cableIsPluggedAfter[Cablestarts,Positions];
	\end{verbatim}
}

Because a cable can either be completely inserted at one position in the permutation or only be inserted with one cable end or not be inserted at all, the following constraints hold:

{\scriptsize
	\begin{verbatim}
	forall(p in Cablestarts, t in Positions) {
	cableIsStoredAtPosition[p,t] + cableIsPluggedBefore[p,t] 
	+ cableIsPluggedAfter[p,t] == 1;
	minimum[p]-t+1 <= bigM*(1-cableIsStoredAtPosition[p,t]);
	t-maximum[p]+1 <= bigM*(1-cableIsStoredAtPosition[p,t]);
	t-minimum[p]   <= bigM*(1-cableIsPluggedBefore[p,t]);
	maximum[p]-t   <= bigM*(1-cableIsPluggedAfter[p,t]);}
	\end{verbatim}
}

The maximum number of cables $M$ that are simultaneously contained in storage is set implicitly by a constraint. $M$ is only set correctly, because it is part of the weighted sum of the four optimization criteria. If not being part of the weighted sum, the value of $M$ would be set arbitrarily large, such that it just satisfies the constraint. Similarly, the value $L$ is also set implicitly by a constraint, whereas the calculation of $N$ and $S$ is straightforward:

{\scriptsize
	\begin{verbatim}
	forall(t in Positions) sum(p in Cablestarts) 
	cableIsStoredAtPosition[p,t] <= M;
	
	forall(p in Cablestarts) maximum[p] - minimum[p] - 1 <= L;
	
	N == sum(s in SoftAtomicConstraints)
	(pfc[s.cafter] - pfc[s.cbefore] <= 0);
	
	S == sum(p in Cablestarts) cableIsStored[p];
	\end{verbatim}
}

The objective function is identical to the one used in the  $\mathbf{M}_\mathbf{C}$ model. 

{\scriptsize
	\begin{verbatim}	  		  	
	minimize S * pow(k, 3)	+ M * pow(k, 2) + L * pow(k, 1) +  N;
	\end{verbatim}
}

For the Gurobi solver, we manually implemented the models $\mathbf{M}_\mathbf{I}$, $\mathbf{M}_\mathbf{B}$, and $\mathbf{M}_\mathbf{B}$  in the 
 C$\sharp$ API of this solver. 
\end{document}